\documentclass[a4paper,fleqn]{cas-dc}

\usepackage[round,authoryear]{natbib}
\usepackage{algorithm}
\usepackage{algorithmic}
\usepackage{tabularx} 
\usepackage{lineno}

\begin{document}
\let\WriteBookmarks\relax
\def\floatpagepagefraction{1}
\def\textpagefraction{.001}

\shorttitle{KA2L: A Knowledge-Aware Active Learning Framework for LLMs}

\shortauthors{H. Yin et al.}

\title [mode = title]{KA2L: A Knowledge-Aware Active Learning Framework for LLMs}    

\affiliation[1]{organization={Faculty of Computing, Harbin Institute of Technology},
    city={Harbin},
    postcode={150001},
    country={China}}

\affiliation[2]{organization={Institute for Advanced Algorithms Research},
    city={Shanghai},
    postcode={201306},
    country={China}}

\affiliation[3]{organization={National Key Laboratory of Smart Farm Technologies and Systems},
    city={Harbin},
    postcode={150001},
    country={China}}

\author[1]{Haoxuan Yin}[orcid=0009-0001-7250-5951]
\ead{25S103196@stu.hit.edu.cn}
\credit{Methodology, Investigation, Software, Formal analysis, Writing - original draft}

\author[2]{Chen Tang}[orcid=0000-0001-7010-2203]
\ead{tangc@iaar.ac.cn}
\credit{Writing - review \& editing}

\author[1]{Yangfan Wang}
\ead{yf.wang@stu.hit.edu.cn}
\credit{Visualization, Writing - review \& editing}

\author[1]{Lian Yan}[orcid=0009-0008-1695-2467]
\ead{23b903008@stu.hit.edu.cn}
\credit{Writing - review \& editing}

\author[1,3]{Jingchi Jiang}[orcid=0000-0003-2167-4082]
\cormark[1] 
\ead{jiangjingchi@hit.edu.cn}
\credit{Conceptualization, Supervision, Resources, Writing - review \& editing}

\cortext[cor1]{Corresponding author}

\begin{abstract}
Fine-tuning large language models (LLMs) with high-quality knowledge has been shown to enhance their performance effectively. However, there is a paucity of research on the depth of domain-specific knowledge comprehension by LLMs and the application of targeted active learning to improve their expertise. To address this gap, we introduce the \textbf{Knowledge-Aware Active Learning (KA2L)} framework. This framework assesses LLMs' mastery of specific knowledge points to aid in constructing unanswerable or unknowable questions through latent space analysis. This active learning strategy enhances training efficiency by focusing on knowledge the model has yet to master, thereby minimizing redundancy in learning already acquired information. This study innovatively employs a knowledge distribution probing technique to examine the hidden states of specific Transformer layers and identify the distribution of known and unknown knowledge within the LLM. Additionally, a hidden-state decoding method is proposed to generate numerous unknown questions in natural language from the latent knowledge space. In our experiments, we selected nine open-source LLMs to validate the effectiveness of the proposed framework. Results indicate that KA2L not only significantly reduces 50\% annotation and computation costs across two open-domain and one vertical-domain dataset but also achieves better performance, offering valuable insights into active learning strategies for LLMs. The code is available at \url{https://github.com/greenjerry/KA2L}.
\end{abstract}



\maketitle

\section{Introduction}

Large Language Models (LLMs) such as GPT-4 and Llama-3 have demonstrated remarkable capabilities across a wide range of NLP tasks \citep{zhao2025surveylargelanguagemodels}, and there is a growing demand for applying them to specific domains \citep{JIANG2025113197}. Enhancing the domain-specific knowledge of LLMs primarily relies on techniques such as Supervised Fine-Tuning (SFT) \citep{zhao2025surveylargelanguagemodels} and Retrieval Augmented Generation (RAG) \citep{gao2024retrievalaugmentedgenerationlargelanguage}. These methods typically require substantial amounts of high-quality annotated data or external knowledge bases. However, in practical applications, two significant challenges arise: (1) The knowledge mastered by LLMs is often invisible, making it necessary to train on the entire domain knowledge during each SFT process, leading to significant resource waste. (2) Without visibility into the model's knowledge, it is difficult to explicitly ascertain the new knowledge required by the LLM. This results in a "black-box" learning process, where incremental learning of new knowledge is prone to noise due to the typically low proportion of new knowledge in the overall training set, ultimately affecting the model's learning efficiency. Therefore, this paper proposes a novel active learning framework that focuses on detecting the distribution of known and unknown knowledge within LLMs. By directing training toward under-learned or unknown knowledge, the framework avoids redundant annotation and repetitive learning on already-acquired concepts, enabling more efficient and targeted knowledge acquisition.

Traditional active learning (AL) aims to identify a small subset of high-value samples from a large data pool, allowing a model to approximate the performance attainable with full-dataset training. Prominent strategies include uncertainty-based, diversity-based, and gradient-based approaches. For instance, diversity-based methods like Coreset \citep{sener2018active} select samples that are maximally different from one another to enhance model generalization. Hybrid approaches such as BADGE \citep{ash2020deep} leverage gradients on models like ResNet \citep{he2016deep}, embodying the core principle of selecting samples based on a combination of uncertainty and diversity. However, directly applying these methods to modern LLMs is challenging due to prohibitive computational costs and a paradigm mismatch, as they were primarily designed for classification rather than generative tasks. Consequently, current AL research for LLM has largely shifted toward distillation-based (e.g. FreeAL \citep{xiao-etal-2023-freeal}) or in-context learning optimization methods \citep{margatina-etal-2023-active}. A key limitation is that they do not assess the model's mastery of learned and to-be-learned content nor analyze the correlation between the model's latent space distribution and the semantic features of upcoming knowledge. This hinders the controllable training of LLMs and the incremental expansion of their unmastered knowledge.

To address these critical bottlenecks, we introduce the Knowledge-Aware Active Learning (KA2L) framework. The core originality of KA2L lies in its system-level paradigm shift, as it conceptually bridges two previously isolated domains: inference-time hallucination detection and training-time Active Learning. Instead of relying on traditional AL heuristics like gradients or representation distance, KA2L pioneers a new paradigm that operationally defines an LLM's "unknown knowledge" through the lens of semantic consistency. Within this paradigm, we operationally define an LLM's "unknown knowledge" as its inability to stably generate semantically consistent answers to a factual question. This phenomenon is quantified by high Semantic Entropy (SE) \citep{Farquhar2024,kuhn2023semantic}, a metric at the heart of our approach. The core objective of this framework is thus to accurately assess this knowledge boundary, thereby efficiently guiding the construction of SFT datasets. Specifically, the KA2L framework employs a Knowledge Distribution Probing mechanism that utilizes hidden states from specific Transformer layers. It performs clustering based on semantic entailment and uses SE to unsupervisedly train a Multi-Layer Perceptron (MLP) as a classifier, which categorizes the question set into "Known" and "Unknown" parts. Furthermore, the KA2L framework proposes a novel application of LLM interpretability techniques for data augmentation. By repurposing "hidden-state decoding" from a mere interpretability tool into an active data generator, KA2L can "reverse engineer" a large volume of natural-sounding question variants from the hidden-space representations corresponding to identified "Unknown" regions. By incorporating these questions into the training data, the KA2L framework establishes an active learning closed loop, enabling the model to concentrate on learning knowledge it has not yet mastered and thereby minimizing repetitive learning and annotation redundancy.

The main contributions of this paper can be summarized as follows:
\begin{enumerate}
\item We propose a novel Knowledge-Aware Active Learning framework (KA2L) that can accurately assess an LLM's degree of knowledge mastery and, by integrating hidden state decoding techniques, actively mine the model's unknown knowledge to guide efficient incremental learning.
\item We innovatively bridge the domains of LLM hallucination detection and active learning. By utilizing Semantic Entropy as an unsupervised signal to probe hidden states, and creatively repurposing hidden-state decoding as a targeted data augmentation engine, we establish a scalable pipeline that offers new avenues for understanding and shaping the internal knowledge boundaries of LLMs.
\item Through extensive experiments on nine open-source LLMs and three datasets, we demonstrate that KA2L not only achieves performance comparable to fine-tuning on the full dataset while reducing annotation and computational costs by approximately $50\%$, but also significantly outperforms adapted classic active learning methods, including Coreset and BADGE, providing a novel and cost-effective solution for fine-tuning LLMs.
\end{enumerate}

\section{Related Work}
\subsection{Active Learning for LLMs}
Active learning is a well-established field for reducing data annotation costs, with classic strategies primarily pivoting on principles of uncertainty, diversity, or a hybrid of both. Prominent methods include uncertainty sampling (e.g., using prediction entropy), diversity-based approaches like Coreset \citep{sener2018active} which selects a representative subset of data, and hybrid methods such as BADGE \citep{ash2020deep} that unify both principles via gradient embeddings. However, transplanting these methods, originally designed for models like CNNs, to modern generative LLMs presents significant challenges. For instance, gradient-based methods like BADGE or Fisher information-based methods like BAIT \citep{ash2021gone} become computationally prohibitive due to the immense scale of LLM parameters, and their core logic does not straightforwardly apply to generative tasks. To our knowledge, systematic adaptation and evaluation of these classic methods for LLM fine-tuning has been limited. In our work, we implement practical adaptations of these strategies to compare with our methods. In addition, active learning theory has also demonstrated efficacy in complex tasks such as meta-reinforcement learning \citep{10.1007/978-3-031-20309-1_31} and causal structure discovery \citep{JIANG2025127466}. However, within the specific context of LLMs, research has focused on alternative goals, such as collaborative learning without human supervision \citep{xiao-etal-2023-freeal}, utilizing knowledge graph feedback to reduce annotation dependence \citep{yan-etal-2025-rlkgf}, or optimizing demonstrations for in-context learning \citep{margatina-etal-2023-active}. In contrast to these approaches, our work introduces a fundamentally different selection signal, semantic entropy, derived from the semantic consistency across multiple model generations. This allows us to directly probe the model's knowledge stability, bypassing the computational hurdles of classic methods while offering a more direct proxy for knowledge gaps in generative tasks.

\subsection{LLM Hallucination Detection}
Given that hallucinations \citep{shuster2021retrievalaugmentationreduceshallucination} are a direct manifestation of an LLM's unknown knowledge, we define the problem of identifying LLM knowledge distribution as a hallucination detection task, also termed output uncertainty quantification \citep{li2023inferencetime}. Current methods fall into two main categories. Output-based methods assess uncertainty from surface features like text or logits. Examples include sampling-based consistency checks \citep{manakul2023selfcheckgpt} and Semantic Entropy (SE) \citep{Farquhar2024,kuhn2023semantic}, which quantifies semantic similarity across multiple outputs to handle linguistic variations. However, these methods often incur high computational costs due to repeated sampling \citep{kossen2024semanticentropyprobesrobust}. In contrast, probing methods leverage internal hidden states, training a classifier to predict output uncertainty \citep{azaria-mitchell-2023-internal,li2023inferencetime}. \citet{azaria-mitchell-2023-internal} demonstrated that hidden states contain veracity signals, enabling efficient detection. More recently, Semantic Entropy Probes (SEP) \citep{kossen2024semanticentropyprobesrobust} were proposed, training a classifier to predict SE values directly from hidden states. By using SE as an unsupervised learning target, SEP achieves performance comparable to surface SE methods but with significantly lower computational cost and improved generalization. The understanding of LLM knowledge boundaries is rapidly evolving, as systematically reviewed in recent comprehensive surveys \citep{li-etal-2025-knowledge-boundary}. Within this broader context, concurrent studies have extended internal-state probing to inference-time applications, such as adaptive retrieval routing based on latent uncertainty \citep{yao-etal-2025-seakr} and online confidence calibration via question reformulation \citep{ni-etal-2025-towards}.

\subsection{LLM Hidden State Decoding}
One of the core ideas of this research is to leverage LLM hidden state decoding techniques to mine potential related knowledge from the internal representations generated when the model processes specific questions (particularly those in the "Unknown" region)\citep{lv2024specfuse,tang2025graphmoe}. This allows for the generation of new questions that are diverse in form yet related in terms of knowledge points, effectively augmenting the set of unknown questions. To elucidate the foundation and existing advancements of the techniques adopted in this study, this section will review relevant hidden state decoding methods. \citet{NostalgebraistLogitLens2020,sakarvadia2023attentionlenstoolmechanistically,belrose2023elicitinglatentpredictionstransformers,pal-etal-2023-future}, based on "lens" methods, proposed an approach for directly decoding hidden states by applying minor transformations and utilizing the model's pre-trained unembedding module to convert hidden states directly into logits. The Patchscope \citep{ghandeharioun2024patchscopes} and SelfIE \citep{chen2024selfie} methods patch hidden states into another model, whose output serves as the decoded result, offering higher readability than lens methods. \citet{morris2023text,morris2024language} introduced the Vec2Text method, which trains a T5 \citep{2020t5} model as a decoder to transform hidden state vectors into natural language text. Experiments demonstrated high accuracy and readability when decoding later-layer hidden state vectors from the LLaMA2 model.

\section{Method}
\label{sec:method}
\subsection{Problem Formulation}

Given the abstract nature of knowledge, this study considers "questions" as external manifestations of knowledge; i.e., a model's ability to correctly answer a question signifies its possession of the knowledge represented by that question. Given an LLM, denoted as $M$, and a set of questions $Q = \{q_1, q_2, \dots, q_n\}$, the knowledge distribution of model $M$ over $Q$, denoted $K_{M,Q}$, is defined as a partition of $Q$ into $(Q_k, Q_{unk})$. Here, $Q_k$ represents the set of questions the model can answer correctly with high confidence, while $Q_{unk}$ represents the set of questions for which the model's answers are uncertain. $Q_{unk}$ will guide knowledge mining, as well as dataset annotation and model fine-tuning in downstream tasks.

To mitigate the risk that $Q_{unk}$ may be too small to effectively support downstream fine-tuning, this study further investigates a question augmentation strategy grounded in the model’s internal representations. For any question $q_i \in Q_{unk}$, its internal hidden states generated by model $M$ during processing are utilized. A new set of questions $\{q_i^{(1)}, q_i^{(2)}, \dots, q_i^{(k_i)}\}$ ($k_i \ge 0$), similar in domains and knowledge points, is then generated through hidden state decoding techniques. The augmented set of unknown questions is represented as $Q_{aug} = \bigcup_{q_i \in Q_{unk}} (\{q_i\} \cup \{q_i^{(1)}, \dots, q_i^{(k_i)}\})$.

Downstream tasks include SFT dataset construction and model fine-tuning. Dataset construction can be defined as creating $\mathcal{D}_{unk} = \{\langle q_i, a_i \rangle | q_i \in Q_{aug}\}$, where $a_i$ is the ground-truth answer to $q_i$. The objective of model fine-tuning is formulated as follows:
\begin{equation}
    \theta_{ft} = \arg \min_{\theta} \frac{1}{N} \sum_{(q_i, a_i) \in \mathcal{D}_{unk}} Loss(M(q_i; \theta), a_i)
\end{equation}

where $\theta_{ft}$ are the fine-tuned model parameters, $\theta$ are the parameters to be optimized during fine-tuning, $N$ is the size of the dataset $\mathcal{D}_{unk}$, $Loss$ is the loss function, and $M$ is the model. Downstream tasks are not the primary focus of this study.

\begin{algorithm}[t]
\caption{Training Process of KA2L Framework}
\label{alg:ka2l}
\begin{algorithmic}[1]
\REQUIRE Large Language Model $\mathcal{M}$, Unlabeled Question Pool $\mathcal{Q}$, Initial Question Set $\mathcal{Q}_{init}$
\ENSURE Fine-tuned Model $\mathcal{M}_{ft}$

\STATE \textbf{Stage 1: Knowledge Distribution Probing}
\FOR{each question $q_i \in \mathcal{Q}_{init}$}
    \STATE Sample hidden states $H_i$ from $\mathcal{M}$
    \STATE Sample multiple outputs $S_i$ and calculate Semantic Entropy (SE)
    \STATE Generate label $y_i$ based on SE thresholding (Eq. 4-5)
\ENDFOR
\STATE Train MLP Classifier $\mathcal{C}$ using $\{(H_i, y_i)\}$

\STATE \textbf{Stage 2: Unknown Knowledge Mining}
\STATE $\mathcal{Q}_{unk} \leftarrow \{q \in \mathcal{Q} \mid \mathcal{C}(\text{GetHidden}(q)) = \text{Unknown}\}$
\STATE Initialize Augmented Set $\mathcal{Q}_{aug} \leftarrow \emptyset$
\FOR{each $q_u \in \mathcal{Q}_{unk}$}
    \STATE Extract hidden state $h_u$
    \STATE Decode $h_u$ into new questions $\{q'_{u}\}$ using Hidden-State Decoder
    \STATE $\mathcal{Q}_{aug} \leftarrow \mathcal{Q}_{aug} \cup \{q_u\} \cup \{q'_{u}\}$
\ENDFOR

\STATE \textbf{Stage 3: Targeted Fine-tuning}
\STATE Annotate $\mathcal{Q}_{aug}$ to get pairs $\{(q, a)\}$
\STATE Fine-tune $\mathcal{M}$ on $\{(q, a)\}$ to obtain $\mathcal{M}_{ft}$ (Eq. 1)

\RETURN $\mathcal{M}_{ft}$
\end{algorithmic}
\end{algorithm}

\subsection{Knowledge Distribution Probing}
\label{sec:method_kdp}

\begin{figure*}[t]
    \centering
    \includegraphics[width=0.98\linewidth]{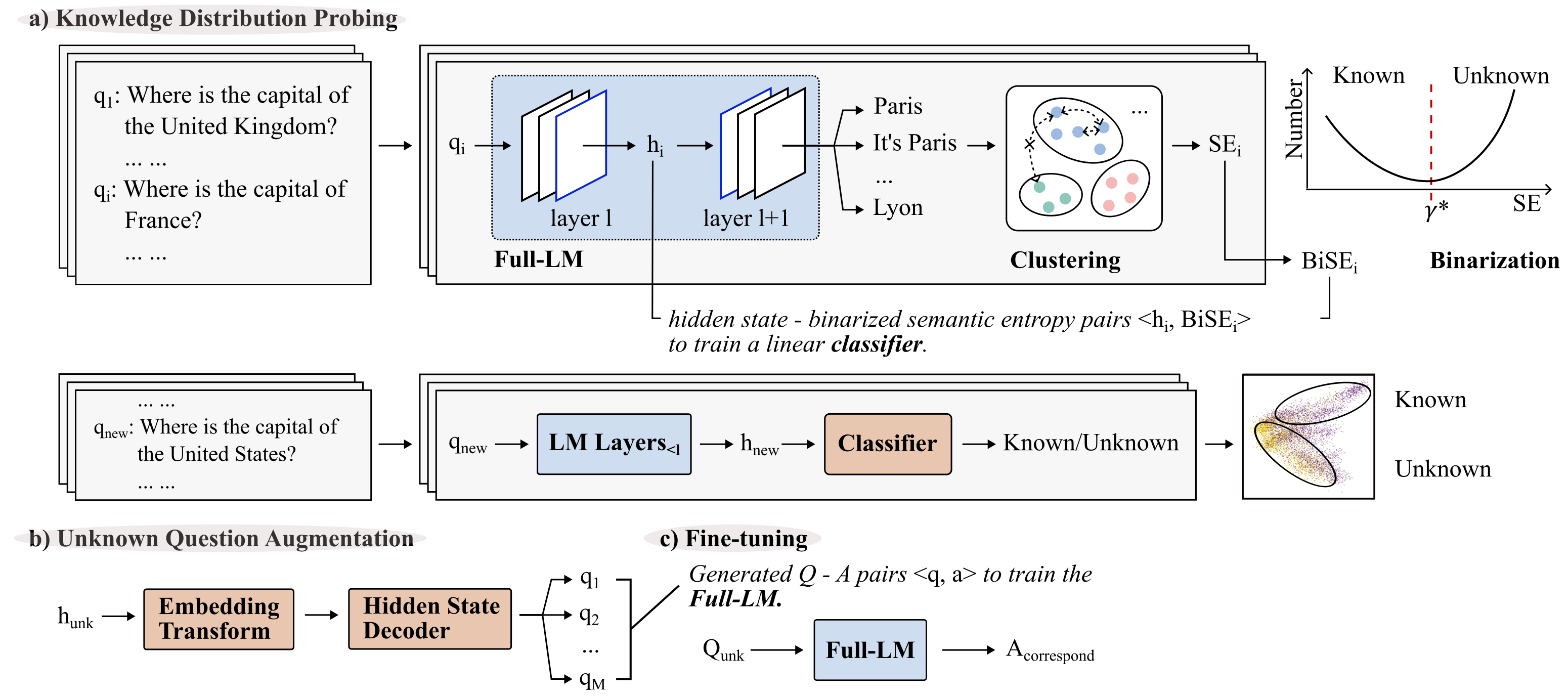}
    \caption{\small{\textbf{KA2L Workflow}:
        \textbf{(a) Knowledge Distribution Probing}:
        \textbf{Training Phase}: For each question in the sampled question set, sample its hidden state once and its textual outputs multiple times. Perform semantic clustering on the textual outputs to calculate Semantic Entropy (SE). The SE is then binarized using a dynamic threshold to serve as labels for the classifier. An MLP classifier is trained using these hidden states and the binarized SE (BiSE).
        \textbf{Inference Phase}: For a new set of questions, sample their $l$-th layer hidden states. These are then classified by the MLP classifier, representing the knowledge distribution as "Known" and "Unknown" knowledge.
        \textbf{(b) Unknown Question Augmentation}: Sample the hidden states ($h_{unk}$) of questions identified as "Unknown" from the knowledge distribution. These are then transformed and decoded into multiple similar questions.
        \textbf{(c) Downstream Tasks}: This knowledge distribution guides dataset construction and model fine-tuning. Many existing methods can be applied, such as LoRA \citep{hu2022lora} and P-tuning \citep{liu-etal-2022-p}.}}
    \label{main-figure}
\end{figure*}

The objective of knowledge distribution probing is to identify the intrinsic distribution of "known" and "unknown" questions for an LLM $M$ over a specific question set $Q$. This task is highly correlated with the goal of LLM hallucination detection. As illustrated in Figure~\ref{main-figure} (a), our probing framework utilizes semantic entropy \citep{Farquhar2024,kuhn2023semantic} as a metric for quantifying uncertainty in model outputs and employs a Multi-Layer Perceptron (MLP) as the classifier to distinguish whether the knowledge corresponding to the hidden state is mastered by the model. The main workflow includes hidden state and model output sampling, semantic entropy calculation and label construction, classifier training, and inference.

\subsubsection{Hidden State and Model Output Sampling}

As shown in Figure~\ref{main-figure} (a), for each question $q_i$ in the question set $Q$, the hidden states $H_i = \{h_i^l\}_{l=1}^L$ of the last token across all $L$ layers are obtained from LLM $M$ at a low temperature (e.g., temperature=$0.1$). Subsequently, for the same question $q_i$, multiple independent samples are drawn at a higher decoding temperature, yielding a set of output sentences $S_i = \{s_i^{(1)}, s_i^{(2)}, \dots, s_i^{(k)}\}$. This process constructs the dataset $\mathcal{S} = \{\langle q_i, H_i, S_i \rangle\}_{q_i \in Q}$ for subsequent classifier training.

\subsubsection{Semantic Entropy Calculation and Label Construction}

The calculation of semantic entropy (SE) follows the methodology proposed in \citet{Farquhar2024,kuhn2023semantic}. It involves performing semantic clustering on the multiple sampled outputs $S_i$ for the same question $q_i$ and quantifying the consistency of the output content based on the clustering results. A lower SE value indicates higher semantic consistency across multiple outputs; conversely, a higher value suggests greater semantic divergence, indicating that the model has not mastered the knowledge associated with the question. Semantic clustering employs a pre-trained Natural Language Inference (NLI) model (e.g., DeBERTa \citep{he2021debertadecodingenhancedbertdisentangled}) to determine semantic equivalence between sentences: if sentence A entails sentence B and sentence B entails sentence A, they are considered semantically equivalent. We deliberately perform clustering on the generated text level via NLI rather than directly on the LLM's latent hidden states. This is because autoregressive hidden states are highly entangled with syntactic noise, word order, and sequence lengths, which can heavily distort semantic boundaries. Utilizing NLI aligns with the standard, mathematically grounded protocols for Semantic Entropy estimation \citep{kuhn2023semantic,Farquhar2024}, ensuring that clustering strictly reflects factual semantic equivalence while filtering out syntactic variations.

For the output set $S_i$ of question $q_i$, let the set of semantic equivalence classes be $\mathcal{C}_{q_i} = \{c_1, c_2, \dots, c_{| \mathcal{C}_{q_i} |}\}$, where $c_k$ is an equivalence class, $|c_k|$ is the number of sentences in that class, and $N = \sum_k |c_k| = |S_i|$ is the total number of samples. The semantic entropy $\mathrm{SE}(S_i)$ can be estimated as:
\begin{equation}
    \mathrm{SE}(S_i) \approx - \sum_{k=1}^{|\mathcal{C}_{q_i}|} \frac{|c_k|}{N} \ln \frac{|c_k|}{N}
\end{equation}
To obtain binary labels (known/unknown) for classifier training, a dynamic thresholding method is applied to binarize the calculated $\mathrm{SE}(S_i)$. Specifically, let $\mathcal{T} = \{\tau_1, \tau_2, \dots, \tau_K\}$ be $K$ candidate thresholds selected within the range of all sample SE values $[\min(\{\mathrm{SE}_i\}_{i=1}^{|Q|}), \max(\{\mathrm{SE}_i\}_{i=1}^{|Q|})]$. For any candidate threshold $\tau \in \mathcal{T}$, each sample's $\mathrm{SE}_i$ is binarized, and the mean-square error (MSE) between its binarized result and the original continuous semantic entropy is calculated:
\begin{equation}
    \mathrm{MSE}(\tau)=\frac{1}{|Q|} \sum_{i=1}^{|Q|} (\mathrm{SE}_i - \mathbf{I}(\mathrm{SE}_i \ge \tau))^2
\end{equation}
where $\mathbf{I}(\cdot)$ is the indicator function, which is 1 if the condition is true and 0 otherwise. Subsequently, the optimal threshold $\gamma^*$ is computed, and $\mathrm{SE}_i$ is binarized to obtain $\mathrm{BiSE}_i$. (See discussion \ref{sec:threshold_robustness} for a detailed robustness analysis demonstrating the effectiveness of this dynamic thresholding method).

\begin{equation}
\label{equ:dynamic_threshold}
    \gamma^* = \arg\min_{\tau \in \mathcal{T}} \mathrm{MSE}(\tau)
\end{equation}
\begin{equation}
    \mathrm{BiSE}_i = \begin{cases} 0 & \text{if } SE_i < \gamma^* \quad (\text{representing known}) \\ 1 & \text{if } SE_i \ge \gamma^* \quad (\text{representing unknown}) \end{cases}
\end{equation}
Finally, the classifier training dataset $\mathcal{D}_{\text{clf}} = \{\langle H_i, \mathrm{BiSE}_i \rangle\}_{q_i \in Q}$ is constructed.

\subsubsection{Classifier Training and Inference}
The dataset $\mathcal{D}_{\text{clf}}$ is partitioned into training, validation, and test sets in a $7$:$2$:$1$ ratio. To effectively discriminate the knowledge states of the LLM, we design a Multi-Layer Perceptron (MLP) as the hidden state classifier. The architectural design of this MLP is informed by efficient MLP components found in modern large language models, such as Llama3, comprising $4$ linear layers and $1$ SiLU activation function. Considering the high dimensionality and potentially complex non-linear features of LLM hidden states, we opted for an MLP structure, aiming for stronger representation learning and pattern recognition capabilities compared to traditional linear classifiers (e.g., logistic regression). The classifier's input dimension matches the hidden state dimension $h_i^l$ of the LLM $M$ under test, the output dimension is $2$, and the intermediate layer dimension is set to $14336$. Training utilizes the standard CrossEntropyLoss function and Adam optimizer \citep{kingma2017adammethodstochasticoptimization}, with a learning rate of $1.0e-5$ for $20$ epochs. For all $L$ hidden layers of each LLM $M$, $L$ independent classifiers are trained. The best-performing classifier $C$ and its corresponding hidden layer number $l$ are selected based on their performance on the test set.

For each question in a new question set $Q_{new}$, we extract the hidden state of the last token at the selected layer $l$. This hidden state is then fed into the trained classifier $C$ for classification, yielding a determination of whether the question is "known" or "unknown," ultimately constructing the knowledge distribution over $Q_{new}$. The classifier $C$ is characterized by its fast operational speed and high parallelizability, rendering its performance overhead on the overall active learning framework negligible. Questions identified as falling within the "Unknown" region of the knowledge distribution are collected and subsequently directed to the hidden state decoding process.

\subsection{Unknown Question Augmentation via Hidden State Decoding}

To further augment the set of questions in the "Unknown" region of the model's knowledge distribution, this study introduces latent space decoding techniques from LLM interpretability research. The advantage of using hidden state decoding for question augmentation is its ability to transcend the surface-level textual limitations of the original "Unknown" questions and mine deeper knowledge associations. The model's hidden states are considered deep, abstract representations of its pre-training data and input information, containing richer semantic and contextual information than the original text \citep{morris2024language,geva-etal-2021-transformer}. By decoding these hidden states, rather than simply paraphrasing the original questions, this method aims to generate new questions that are highly relevant in terms of knowledge points and domain but are more diverse in form. These questions represent the same knowledge point from different perspectives, thereby more comprehensively identifying the model's knowledge deficiencies.
 
To achieve this objective, this study adapts the vec2text method from \citet{morris2024language}, training a decoder from hidden states to text for each model on each dataset, as shown in Figure~\ref{main-figure} (b). The base model for the decoder is t5-base \citep{2020t5}, which is an encoder-decoder architecture Transformer model. The training process employed end-to-end training, simultaneously training the weights of the transformation linear layers and t5 (see section \ref{sec:hidden-state-decoder-detailed-impl} for detailed training dataset composition and decoder parameters), enabling the decoder to learn the ability to reconstruct the original text from the source model's hidden states. The decoding process is as follows: the hidden state $h_{unk}$ from the "Unknown" region of the knowledge distribution is input; it is transformed into t5's hidden state representation via two Linear layers and a GeLU layer, with Gaussian noise added; and then the trained t5-base model is used for inference to obtain natural language text results, thereby augmenting the "Unknown" region of the knowledge distribution.

\subsection{Downstream Tasks}

Downstream tasks include constructing question-answer pairs based on the augmented unknown question set $Q_{aug}$, as well as model fine-tuning and evaluation, as shown in Figure~\ref{main-figure} (c). Downstream tasks are not the primary focus of this paper. The methods are diverse; for example, constructing question-answer pairs can involve methods such as knowledge distillation \citep{xu2024surveyknowledgedistillationlarge}, web search, and manual annotation, which ensures that each newly generated question variant is paired with an independently verified, factually correct answer. Meanwhile, model fine-tuning can utilize techniques like LoRA \citep{hu2022lora} and P-tuning \citep{liu-etal-2022-p}.

\section{Experimental Setup}

To rigorously evaluate the effectiveness and efficiency of our proposed KA2L framework, we conducted a series of experiments. This section details the datasets, models, experimental design, and implementation specifics.

\subsection{Models, Datasets, and Evaluation Metrics}

The experiments selected nine open-source large language models, including Llama \citep{touvron2023llamaopenefficientfoundation}, Mistral \citep{jiang2023mistral7b}, Phi \citep{abdin2024phi3technicalreporthighly}, Qwen \citep{qwen2025qwen25technicalreport}, and GLM \citep{glm2024chatglm}, covering different model architectures and parameter scales. 

To ensure the breadth of the evaluation, the experiments utilized two open-domain question-answering datasets, TriviaQA \citep{joshi-etal-2017-triviaqa} and NQ\_Open \citep{lee-etal-2019-latent}, and one medical domain dataset, MedMCQA \citep{pmlr-v174-pal22a}. These datasets cover different knowledge domains and question types. 

Evaluation metrics including BLEU \citep{papineni-etal-2002-bleu}, ROUGE \citep{lin-2004-rouge}, METEOR \citep{lavie-agarwal-2007-meteor}, and BERTScore \citep{Zhang*2020BERTScore:}, were adopted to comprehensively assess the fluency, accuracy, and semantic similarity of the generated answers.

It is important to note that all our experiments are conducted in a closed-book setting. This means the models rely solely on their internal, parametric knowledge to answer questions, without access to any external information retrieval system during inference.

\subsection{Experimental Design and Data Construction}
\label{sec:exp_design}

The core of our experimental design involves training a knowledge distribution probe for each LLM-dataset pair on a held-out data sample. This trained probe then partitions a separate, larger data pool to construct our primary fine-tuning datasets:

\begin{description}
  \item[\textbf{$D_{\text{unk}}$}] Contains questions identified by the probe as "unknown". This set represents the high-value data actively selected by our KA2L framework.
  \item[\textbf{$D_{\text{k}}$}] Contains questions identified as "known", serving as a baseline to evaluate the utility of data already mastered by the model.
  \item[\textbf{$D_{\text{combine}}$}] A balanced mix of samples from $D_{\text{unk}}$ and $D_{\text{k}}$, simulating a standard, unfiltered dataset collected without an active learning strategy.
\end{description}

Based on these partitions, we constructed three specific experimental sets for fine-tuning comparisons:

\begin{description}
    \item \textbf{10k Unknown:} Consists of $10,000$ selected unknown samples from $D_{\text{unk}}$.
    \item \textbf{5k Unknown:} A random $5,000$-sample subset of the 10k Unknown set.
    \item \textbf{10k Combine:} Constructed by mixing $5,000$ unknown samples with $5,000$ known samples.
\end{description}

Additionally, to evaluate our data augmentation module, we created a \textbf{10k Augmented} dataset by generating $5,000$ new questions from the hidden states of the \textit{5k Unknown} set and combining them with the original samples.

\subsection{Implementation Details}
All experiments were conducted on NVIDIA A100 40GB GPUs. The implementation of KA2L consists of the following three key components.

\subsubsection{Knowledge Distribution Probe} 
For each model, we trained a Multi-Layer Perceptron (MLP) as the probe. As detailed in Section \ref{sec:method_kdp}, the probe takes the hidden states of a specific layer as input. To determine the optimal layer for each model, we conducted a layer-wise performance analysis (comprehensive results are discussed in Section \ref{sec:layer-wise-analysis}). Based on this analysis, we selected the layer with the highest probing accuracy (e.g., layer 31 for Llama-3-8B) for the final experiments (see Table \ref{tab:layer_index_choices}), which remains remarkably consistent across both open and vertical domains and indicates the structural robustness of our probing mechanism. The MLP was trained for 20 epochs with a learning rate of $1\times 10^{-5}$.

\begin{table*}[t]
\centering
\caption{Model-specific layer index choices based on probing accuracy.} 
\label{tab:layer_index_choices}
\begin{tabular}{@{}l c c c@{}}
\toprule
\multirow{2}{*}{\textbf{Model}} & \multicolumn{3}{c}{\textbf{Hidden-State Layer Index}} \\
\cmidrule(lr){2-4}
& \textbf{TriviaQA Mix} & \textbf{NQ-Open Mix} & \textbf{MedMCQA Mix} \\
\midrule
DeepSeek-R1-Distill-Qwen-7B & 28 & 28 & 28 \\
glm4-9b-chat & 38 & 38 & 24 \\
Llama-2-7b-chat-hf & 31 & 30 & 30 \\
Llama-3.1-8B-Instruct	 & 31 & 32 & 31 \\
Mistral-7B-Instruct-v0.1 & 31 & 31 & 30 \\
Mistral-7B-Instruct-v0.3 & 31 & 31 & 31 \\
Phi-3.5-mini-instruct & 32 & 32 & 32 \\
Qwen1.5-7B-Chat & 32 & 26 & 19 \\
Qwen2.5-7B-Instruct & 22 & 19 & 28 \\
\bottomrule
\end{tabular}
\end{table*}

\subsubsection{Hidden-State Decoder}
\label{sec:hidden-state-decoder-detailed-impl}
For the data augmentation module, we employed a T5-base model \footnote{\url{https://hf-mirror.com/google-t5/t5-base}} \citep{2020t5} as the decoder.

\textbf{Dataset}: We used $250,000$ data entries for training. Among these, $200,000$ entries are fixed, sampled from one-million-instructions \citep{morris2024language}, to ensure its capability in decoding fundamental questions. An additional $50,000$ entries originate from the TriviaQA \citep{joshi-etal-2017-triviaqa}, NQ\_Open \citep{lee-etal-2019-latent}, and MedMCQA \citep{pmlr-v174-pal22a} datasets used in the experiments of this paper. This subset of data was not used in any other experiments and was solely dedicated to training the decoder for each model on its corresponding dataset. This specific data composition was chosen to balance computational efficiency with effective domain adaptation, ensuring the decoder acquires a sufficient linguistic foundation while strictly preventing any data leakage into the downstream active learning and evaluation sets.

\textbf{Training Parameters}: Given the heterogeneous latent spaces across different LLMs, a separate T5-base decoder is trained independently for each specific LLM-dataset pair. All experiments utilized a fully fine-tuned T5-base model as the decoder. The training involved $40$ epochs, a learning rate of $2.0 \times 10^{-4}$, and $100,000$ warmup steps. The dataset for training the decoder was selected based on the model being decoded and the specific dataset context. The layer numbers for hidden state extraction, presented in Table \ref{tab:layer_index_choices}, were not chosen arbitrarily. Instead, they were empirically determined for each model-dataset pair by selecting the layer that yielded the highest classifier performance (AUROC) in our comprehensive layer-wise analysis (discussed in Section \ref{sec:layer-wise-analysis}).

\textbf{Decoding Process}: The decoding process leverages the trained T5-base to generate new questions. Specifically, a hidden state $h_{\text{unk}}$ from the "Unknown" region is first transformed into the T5's latent space via two linear layers and a GeLU activation function, with Gaussian noise added to promote diversity. The resulting vector is then fed into the trained T5-base model, which generates the new question in natural language.

\subsubsection{Supervised Fine-Tuning (SFT).}
We employed the LLaMA-Factory framework \citep{zheng2024llamafactory} to fine-tune all models. The fine-tuning parameters adopted recommended values to simulate real-world fine-tuning scenarios. A total of $9$ models were fine-tuned $5$ times each across $3$ datasets, resulting in $135$ fine-tuning runs. For a given model, the only parameter difference when fine-tuning on different datasets was the dataset itself; all other parameters remained consistent.

All fine-tuning was performed using LoRA \citep{hu2022lora}, with \texttt{lora\_target} set to "all" for all modules. Bf16, \texttt{FlashAttention-2} \citep{dao2023flashattention2}, and \texttt{fast\_tokenizer} were used to accelerate the fine-tuning process. The number of epochs was set to 3 for all runs. Other parameters are detailed in Table \ref{tab:sft_training_params} below.

For generating outputs from the fine-tuned models during the evaluation phase, a consistent decoding strategy was employed. We used a low temperature of $0.1$ for sampling. This encourages the model to produce more factual and deterministic outputs by reducing randomness, which is appropriate for the question-answering tasks in our evaluation. Other decoding parameters, such as top-p and top-k, were kept at their default values.

\begin{table*}[t]
\small
\centering
\caption{SFT training parameters for different models.}
\label{tab:sft_training_params}
\begin{tabular}{@{}l c c c c}
\toprule
\multirow{2}{*}{\textbf{Model}} & \multirow{2}{*}{\textbf{Learning Rate}} & \multirow{2}{*}{\textbf{Template}} & \multirow{2}{*}{\textbf{Batch Size}} & \multirow{2}{*}{\textbf{\shortstack{Gradient Accumu-\\lation Steps}}} \\ \\
\midrule
\texttt{DeepSeek-R1-Distill-Qwen-7B} & 5.0e-5 & \texttt{qwen} & 4 & 4 \\
\texttt{glm4-9b-chat} & 1.0e-4 & \texttt{glm4} & 2 & 8 \\
\texttt{Llama-2-7b-chat-hf} & 1.0e-4 & \texttt{llama2} & 4 & 2 \\
\texttt{Llama-3.1-8B-Instruct} & 1.0e-4 & \texttt{llama3} & 4 & 2 \\
\texttt{Mistral-7B-Instruct-v0.1} & 1.0e-4 & \texttt{mistral} & 4 & 4 \\
\texttt{Mistral-7B-Instruct-v0.3} & 1.0e-4 & \texttt{mistral} & 4 & 4 \\
\texttt{Phi-3.5-mini-instruct} & 1.0e-4 & \texttt{phi} & 4 & 8 \\
\texttt{Qwen1.5-7B-Chat} & 5.0e-5 & \texttt{qwen} & 4 & 4 \\
\texttt{Qwen2.5-7B-Instruct} & 5.0e-5 & \texttt{qwen} & 4 & 4 \\
\bottomrule
\end{tabular}
\end{table*}

\subsection{Research Questions}
Guided by the experimental design above, this study aims to answer the following three research questions:

\paragraph{RQ1: Cost-Efficiency.} Can KA2L achieve comparable performance to a full, unfiltered dataset while using only a fraction (e.g., $50\%$) of the annotation and computational budget? We answer this by comparing fine-tuning on \textit{5k Unknown} data against the \textit{10k Combine} baseline.

\paragraph{RQ2: Selection Effectiveness.} Given an identical data budget, does KA2L's strategy of selecting "unknown" data yield superior performance compared to a naive, unfiltered approach? We investigate this by comparing \textit{10k Unknown} against \textit{10k Combine}.

\paragraph{RQ3: Augmentation Utility.} In scenarios with a limited pool of original "unknown" data, can our hidden-state decoding method effectively augment the training set to further boost performance? We evaluate this by comparing the performance of the \textit{10k Augmented} set against the \textit{5k Unknown} and \textit{10k Unknown} sets.

\subsection{Comparison with Adapted Traditional Active Learning Methods}

To situate KA2L in the broader research context, we performed a comparative analysis against several traditional active learning methods. This comparative experiment was performed on the LLaMA-3.1-8B-Instruct model across all three datasets, where each method was tasked with selecting a $5,000$-sample subset from the \textit{$10k$ Combine} set. 

Since methods such as Entropy \citep{6889457}, Coreset \citep{sener2018active}, and BADGE \citep{ash2020deep} were originally designed for classification tasks, not for generative LLMs, it is necessary to develop practical adaptations for them, primarily by using prediction entropy as a proxy for uncertainty and final-layer hidden states as embeddings for diversity. Our full adaptation methodology for these strategies is as follows.

Our adaptation strategy centers on a single, efficient pass over the unlabeled data pool to pre-calculate two key metrics for each prompt: an uncertainty score and a diversity embedding. This pre-computation, while resource-intensive, is performed only once, and its results are reused across all comparative methods, ensuring both efficiency and a fair comparison.

\textbf{Uncertainty Score.} Traditional gradient-based uncertainty metrics are infeasible for LLMs. We instead use prediction entropy as a computationally efficient and effective proxy. Specifically, for each prompt, we perform a standard forward pass (temperature=$1.0$) to obtain the next-token logits and calculate the entropy of the resulting probability distribution, $H(p) = -\sum_i p_i \log p_i$. This captures the model's intrinsic confidence and serves as the uncertainty score.

\textbf{Diversity Embedding.} To measure diversity, we leverage the semantic representation of each prompt from the LLM itself. Instead of computationally prohibitive gradient embeddings (as in the original BADGE), we extract the hidden-state representation of the final prompt token from a deep model layer (see Table \ref{tab:layer_index_choices}). This high-dimensional vector captures the prompt's semantic essence, enabling a meaningful measure of diversity.

Using these two pre-computed metrics, we implement the comparative strategies as follows:

\textbf{Random Sampling}: Uniformly samples a subset of data from the pool.

\textbf{Uncertainty Sampling}: Selects samples with the highest prediction entropy scores.

\textbf{Coreset Sampling}: This diversity-focused method is implemented using a standard k-Center-Greedy algorithm on the hidden-state embeddings, iteratively selecting the data point furthest from its nearest neighbor in the already selected set.

\textbf{BADGE (Adapted)}: Our adaptation preserves BADGE's hybrid principle by using a weighted k-MEANS++ seeding procedure. The probability of selecting a new sample is proportional to its squared distance to the nearest selected center, multiplied by its uncertainty score (entropy), balancing diversity and uncertainty.

\section{Experimental Results}

In this section, we present the main experimental results to answer the proposed research questions. We first analyze the cost-efficiency and selection effectiveness of KA2L across three datasets (RQ1 and RQ2). Next, we evaluate the utility of our hidden-state decoding augmentation strategy (RQ3). Finally, we compare KA2L with adapted traditional active learning methods to demonstrate its superiority in the context of LLM fine-tuning.

\subsection{KA2L-Guided Fine-tuning Achieves Superior Cost-Efficiency and Effectiveness}

To validate the cost-efficiency (RQ1) and selection effectiveness (RQ2) of our framework, we conducted extensive experiments on MedMCQA, NQ-Open, and TriviaQA. The main results are summarized in Table \ref{tab:medmcqa_results}, Table \ref{tab:nq_results}, and Table \ref{tab:triviaqa_results}, respectively.

\paragraph{Cost-Efficiency (RQ1).} 
A central finding is that fine-tuning with \textit{$5k$ Unknown} samples, actively selected by KA2L, achieves performance comparable to the \textit{$10k$ Combine} setting while using only half the data.
For instance, on the MedMCQA dataset (Table \ref{tab:medmcqa_results}), \texttt{Llama-3.1-8B-Instruct} trained on \textit{5k Unknown} data reaches a ROUGE-L score of 29.96, which is nearly identical to the 30.17 achieved with the full \textit{10k Combine} set. Similarly, on the open-domain NQ-Open dataset (Table \ref{tab:nq_results}), \texttt{glm4-9b-chat} achieves 45.91 ROUGE-L on \textit{5k Unknown}, effectively matching the 45.94 of the \textit{10k Combine} baseline. 
This pattern consistently validates RQ1: KA2L can effectively cut annotation and computational costs by approximately 50\% while maintaining high performance, confirming the high density of information within the selected "unknown" samples.

\paragraph{Selection Effectiveness (RQ2).} 
When comparing datasets of the same size, the superiority of KA2L's targeted selection becomes evident. The \textit{10k Unknown} setting consistently outperforms the \textit{10k Combine} baseline across almost all models and metrics. 
Taking TriviaQA (Table \ref{tab:triviaqa_results}) as an example, \texttt{DeepSeek-R1-Distill-Qwen-7B} shows a significant leap from 23.65 (\textit{10k Combine}) to 27.00 (\textit{10k Unknown}) in ROUGE-L. On MedMCQA, \texttt{Mistral-7B-Instruct-v0.3} improves from 27.55 to 35.28. 
These results provide a strong affirmative answer to RQ2, illustrating that for a fixed budget, intelligently selecting data that addresses a model's specific knowledge gaps is far more effective than an indiscriminate, unfiltered training approach. It also implies that fine-tuning on "known" data offers diminishing returns and can be considered redundant.

\begin{table*}
  \caption{\small{\textbf{Active learning performance on MedMCQA.} KA2L-selected data (Unknown) is compared against Known data and a mixed Combine setting. Bold values indicate the best performance.
  }}
  \label{tab:medmcqa_results}
  \centering
  \scriptsize 
  \renewcommand{\arraystretch}{0.85} 
  \setlength{\tabcolsep}{3pt}
  
  \begin{tabular}{@{}l lcccc|lcccc@{}}
    \toprule
    \textbf{Model} & \textbf{SFT Dataset} & \textbf{BLEU} & \textbf{ROUGE-L} & \textbf{METEOR} & \textbf{BS(\%)} 
    & \textbf{SFT Dataset} & \textbf{BLEU} & \textbf{ROUGE-L} & \textbf{METEOR} & \textbf{BS(\%)} \\
    \midrule
    
    \multirow{3}{*}{\shortstack[l]{DeepSeek-R1-\\Distill-Qwen-7B}} 
    & None          & 0.02 & 0.52  & 1.31 & 76.29 
    & 10k Combine   & 2.44 & 12.55 & 7.92 & 83.70 \\ 
    & 5k Known      & 1.85 & 10.41 & 6.50 & 83.19 
    & 10k Unknown   & \textbf{2.85} & \textbf{14.53} & \textbf{9.20} & \textbf{84.00} \\ 
    & 5k Unknown    & 1.91 & 11.99 & 7.54 & 83.62 
    & 10k Augmented & 2.62 & 13.55 & 8.91 & \textbf{84.00} \\ 
    \midrule
    
    \multirow{3}{*}{\shortstack[l]{glm4-9b-chat}} 
    & None          & 0.08 & 1.71  & 4.21 & 78.48 
    & 10k Combine   & 6.90 & 28.27 & 19.67& 86.80 \\
    & 5k Known      & 4.17 & 20.62 & 13.86& 85.17 
    & 10k Unknown   & \textbf{9.82} & \textbf{36.02} & \textbf{25.50}& \textbf{88.30} \\
    & 5k Unknown    & 6.61 & 28.44 & 19.87& 86.73 
    & 10k Augmented & 8.48 & 29.92 & 21.67& 87.13 \\
    \midrule

    \multirow{3}{*}{\shortstack[l]{Llama-2-7b-\\chat-hf}}
    & None          & 0.06 & 0.99  & 2.54 & 77.33 
    & 10k Combine   & 5.37 & 23.76 & 15.80& 85.90 \\
    & 5k Known      & 3.34 & 16.13 & 10.45& 84.35 
    & 10k Unknown   & \textbf{7.99} & \textbf{29.78} & \textbf{20.14}& \textbf{87.15} \\
    & 5k Unknown    & 5.91 & 23.25 & 15.54& 85.79 
    & 10k Augmented & 5.56 & 23.77 & 16.06& 85.92 \\
    \midrule

    \multirow{3}{*}{\shortstack[l]{Llama-3.1-8B-\\Instruct}}
    & None          & 0.10 & 2.34  & 5.01 & 78.58 
    & 10k Combine   & 8.21 & 30.17 & 21.11& 87.14 \\
    & 5k Known      & 5.72 & 23.29 & 16.18& 85.71 
    & 10k Unknown   & \textbf{10.82}& \textbf{36.55} & \textbf{25.97}& \textbf{88.49} \\
    & 5k Unknown    & 8.49 & 29.96 & 20.84& 87.20 
    & 10k Augmented & 8.81 & 30.54 & 21.36& 87.31 \\
    \midrule

    \multirow{3}{*}{\shortstack[l]{Mistral-7B-\\Instruct-v0.1}}
    & None          & 0.12 & 3.17  & 6.29 & 79.60 
    & 10k Combine   & 7.06 & 27.86 & 19.16& 86.70 \\
    & 5k Known      & 3.35 & 18.09 & 12.03& 84.74 
    & 10k Unknown   & \textbf{9.71} & \textbf{36.11} & \textbf{25.20}& \textbf{88.37} \\
    & 5k Unknown    & 6.97 & 27.33 & 18.82& 86.65 
    & 10k Augmented & 7.43 & 28.85 & 20.29& 86.92 \\
    \midrule

    \multirow{3}{*}{\shortstack[l]{Mistral-7B-\\Instruct-v0.3}}
    & None          & 0.11 & 2.10  & 5.19 & 79.22 
    & 10k Combine   & 6.88 & 27.55 & 19.29& 86.76 \\
    & 5k Known      & 4.32 & 19.17 & 13.11& 85.03 
    & 10k Unknown   & \textbf{9.58} & \textbf{35.28} & \textbf{25.08}& \textbf{88.30} \\
    & 5k Unknown    & 6.71 & 27.02 & 18.79& 86.63 
    & 10k Augmented & 8.24 & 28.94 & 20.86& 87.07 \\
    \midrule

    \multirow{3}{*}{\shortstack[l]{Phi-3.5-mini-\\instruct}}
    & None          & 0.09 & 1.59  & 4.10 & 78.44 
    & 10k Combine   & 7.42 & 27.70 & 18.57& 86.74 \\
    & 5k Known      & 6.63 & 25.40 & 16.90& 86.30 
    & 10k Unknown   & 8.05 & \textbf{29.69} & 19.95& \textbf{87.21} \\
    & 5k Unknown    & 7.61 & 27.85 & 18.59& 86.81 
    & 10k Augmented & \textbf{8.30} & 29.18 & \textbf{20.05}& 87.07 \\
    \midrule

    \multirow{3}{*}{\shortstack[l]{Qwen1.5-7B-\\Chat}}
    & None          & 0.08 & 1.64  & 4.03 & 78.63 
    & 10k Combine   & 4.60 & 19.65 & 13.12& 85.00 \\
    & 5k Known      & 3.11 & 15.09 & 9.95 & 84.10 
    & 10k Unknown   & \textbf{5.46} & \textbf{23.72} & \textbf{15.78}& \textbf{85.84} \\
    & 5k Unknown    & 4.13 & 19.34 & 12.86& 84.90 
    & 10k Augmented & 4.85 & 21.55 & 14.84& 85.40 \\
    \midrule

    \multirow{3}{*}{\shortstack[l]{Qwen2.5-7B-\\Instruct}}
    & None          & 0.09 & 1.80  & 4.34 & 78.35 
    & 10k Combine   & 6.30 & 24.13 & 16.80& 85.93 \\
    & 5k Known      & 4.67 & 20.30 & 13.85& 85.06 
    & 10k Unknown   & 6.63 & \textbf{27.16} & \textbf{18.79}& \textbf{86.54} \\
    & 5k Unknown    & \textbf{6.80} & 24.07 & 16.58& 85.95 
    & 10k Augmented & 5.92 & 24.65 & 17.73& 86.14 \\
    
    \bottomrule
  \end{tabular}
\end{table*}

\begin{table*}
  \caption{\small{\textbf{Active learning performance on the NQ\_Open dataset.} Bold values indicate the best performance.}}
  \label{tab:nq_results}
  \centering
  \scriptsize 
  \renewcommand{\arraystretch}{0.95} 
  \setlength{\tabcolsep}{3pt}      
  
  \begin{tabular}{@{}l lcccc|lcccc@{}}
    \toprule
    \textbf{Model} & \textbf{SFT Dataset} & \textbf{BLEU} & \textbf{ROUGE-L} & \textbf{METEOR} & \textbf{BS(\%)} 
    & \textbf{SFT Dataset} & \textbf{BLEU} & \textbf{ROUGE-L} & \textbf{METEOR} & \textbf{BS(\%)} \\
    \midrule
    
    \multirow{3}{*}{\shortstack[l]{DeepSeek-R1-\\Distill-Qwen-7B}} 
    & None          & 0.02 & 0.46  & 1.26 & 76.39 
    & 10k Combine   & 6.10 & 13.17  & 9.68 & 86.45 \\ 
    & 5k Known      & 4.69 & 11.15  & 7.87 & 86.19 
    & 10k Unknown   & \textbf{7.48} & \textbf{16.05} & \textbf{11.84} & \textbf{86.98} \\ 
    & 5k Unknown    & 4.05 & 12.19  & 8.78 & 86.37 
    & 10k Augmented & 2.85 & 13.56 & 10.04 & 86.49 \\ 
    \midrule
    
    \multirow{3}{*}{\shortstack[l]{glm4-9b-chat}} 
    & None          & 0.22 & 3.68  & 8.63 & 79.94 
    & 10k Combine   & 28.78 & 45.94  & 36.50 & 91.33 \\ 
    & 5k Known      & 19.55 & 35.78  & 27.67 & 89.62 
    & 10k Unknown   & \textbf{39.54} & \textbf{57.89} & \textbf{46.91} & \textbf{93.27} \\ 
    & 5k Unknown    & 28.61 & 45.91  & 36.47 & 91.31 
    & 10k Augmented & 22.49 & 46.00 & 36.62 & 91.34 \\ 
    \midrule

    \multirow{3}{*}{\shortstack[l]{Llama-2-7b-\\chat-hf}} 
    & None          & 0.12 & 1.92  & 4.70 & 78.66 
    & 10k Combine   & 21.29 & 39.10  & 31.04 & 90.25 \\ 
    & 5k Known      & 12.23 & 25.65  & 19.71 & 88.06 
    & 10k Unknown   & \textbf{33.56} & \textbf{52.14} & \textbf{41.94} & \textbf{92.45} \\ 
    & 5k Unknown    & 23.47 & 39.25  & 31.15 & 90.23 
    & 10k Augmented & 17.78 & 38.69 & 30.72 & 90.24 \\ 
    \midrule

    \multirow{3}{*}{\shortstack[l]{Llama-3.1-8B-\\Instruct}} 
    & None          & 0.15 & 3.49  & 7.93 & 79.12 
    & 10k Combine   & 24.46 & 45.15  & 35.35 & 91.09 \\ 
    & 5k Known      & 16.57 & 32.56  & 24.69 & 89.07 
    & 10k Unknown   & \textbf{38.91} & \textbf{58.84} & \textbf{46.97} & \textbf{93.35} \\ 
    & 5k Unknown    & 22.32 & 44.94  & 35.04 & 91.08 
    & 10k Augmented & 21.90 & 44.29 & 34.53 & 91.03 \\ 
    \midrule

    \multirow{3}{*}{\shortstack[l]{Mistral-7B-\\Instruct-v0.1}} 
    & None          & 0.31 & 5.44  & 10.85 & 80.48 
    & 10k Combine   & 21.21 & 35.27  & 27.16 & 89.76 \\ 
    & 5k Known      & 13.35 & 26.88  & 20.29 & 88.39 
    & 10k Unknown   & \textbf{38.70} & \textbf{58.51} & \textbf{46.23} & \textbf{93.49} \\ 
    & 5k Unknown    & 20.44 & 34.59  & 26.69 & 89.56 
    & 10k Augmented & 14.05 & 35.07 & 27.20 & 89.70 \\ 
    \midrule

    \multirow{3}{*}{\shortstack[l]{Mistral-7B-\\Instruct-v0.3}} 
    & None          & 0.28 & 3.26  & 8.21 & 79.99 
    & 10k Combine   & 28.64 & 46.53  & 36.27 & 91.37 \\ 
    & 5k Known      & 16.17 & 33.89  & 25.37 & 89.31 
    & 10k Unknown   & \textbf{39.45} & \textbf{59.72} & \textbf{47.23} & \textbf{93.65} \\ 
    & 5k Unknown    & 28.30 & 46.28  & 35.99 & 91.44 
    & 10k Augmented & 24.32 & 46.99 & 36.72 & 91.51 \\ 
    \midrule

    \multirow{3}{*}{\shortstack[l]{Phi-3.5-mini-\\instruct}} 
    & None          & 0.15 & 2.11  & 5.42 & 79.18 
    & 10k Combine   & 14.30 & 27.85  & 20.88 & 88.52 \\ 
    & 5k Known      & 12.83 & 24.73  & 18.19 & 88.03 
    & 10k Unknown   & \textbf{20.01} & \textbf{34.98} & \textbf{26.71} & \textbf{89.66} \\ 
    & 5k Unknown    & 14.28 & 27.83  & 20.87 & 88.55 
    & 10k Augmented & 12.57 & 28.59 & 21.55 & 88.57 \\ 
    \midrule

    \multirow{3}{*}{\shortstack[l]{Qwen1.5-7B-\\Chat}} 
    & None          & 0.20 & 2.72  & 6.51 & 79.55 
    & 10k Combine   & 14.97 & 31.62  & 24.45 & 88.89 \\ 
    & 5k Known      & 10.12 & 22.56  & 16.98 & 87.31 
    & 10k Unknown   & \textbf{23.39} & \textbf{39.90} & \textbf{31.31} & \textbf{90.21} \\ 
    & 5k Unknown    & 17.05 & 30.86  & 23.66 & 88.78 
    & 10k Augmented & 12.98 & 33.72 & 26.53 & 89.28 \\ 
    \midrule

    \multirow{3}{*}{\shortstack[l]{Qwen2.5-7B-\\Instruct}} 
    & None          & 0.21 & 3.43  & 7.99 & 79.73 
    & 10k Combine   & 16.23 & 29.57  & 23.12 & 88.34 \\ 
    & 5k Known      & 9.81 & 24.26  & 18.68 & 87.51 
    & 10k Unknown   & \textbf{17.42} & \textbf{34.76} & \textbf{27.37} & \textbf{89.25} \\
    & 5k Unknown    & 15.90 & 29.91  & 23.32 & 88.41 
    & 10k Augmented & 12.79 & 31.67 & 25.10 & 88.70 \\ 
    \bottomrule
  \end{tabular}
\end{table*}

\begin{table*} 
  \caption{\small{\textbf{Active learning performance on the TriviaQA dataset.} Bold values indicate the best performance.}
  }
  \label{tab:triviaqa_results}
  \centering
  \scriptsize 
  \renewcommand{\arraystretch}{0.95}
  \setlength{\tabcolsep}{3pt}
  
  \begin{tabular}{@{}l lcccc|lcccc@{}}
    \toprule
    \textbf{Model} & \textbf{SFT Dataset} & \textbf{BLEU} & \textbf{ROUGE-L} & \textbf{METEOR} & \textbf{BS(\%)} 
    & \textbf{SFT Dataset} & \textbf{BLEU} & \textbf{ROUGE-L} & \textbf{METEOR} & \textbf{BS(\%)} \\
    \midrule
    
    \multirow{3}{*}{\shortstack[l]{DeepSeek-R1-\\Distill-Qwen-7B}} 
    & None          & 0.02 & 0.49  & 1.24 & 76.13 
    & 10k Combine   & 9.98 & 23.65  & 16.38 & 87.05 \\ 
    & 5k Known      & 7.79 & 18.77  & 12.82 & 86.32 
    & 10k Unknown   & \textbf{14.20} & \textbf{27.00} & \textbf{18.86} & \textbf{87.56} \\ 
    & 5k Unknown    & 9.41 & 22.22  & 15.39 & 86.85 
    & 10k Augmented & 6.68 & 24.46 & 17.11 & 87.16 \\ 
    \midrule
    
    \multirow{3}{*}{\shortstack[l]{glm4-9b-chat}} 
    & None          & 0.15 & 3.62  & 8.16 & 79.13 
    & 10k Combine   & \textbf{21.51} & 41.26  & 30.71 & 88.18 \\ 
    & 5k Known      & 14.68 & 31.65  & 23.03 & 86.34 
    & 10k Unknown   & 11.68 & \textbf{55.80} & \textbf{42.43} & \textbf{91.00} \\ 
    & 5k Unknown    & 19.62 & 40.28  & 29.85 & 87.97 
    & 10k Augmented & 15.21 & 40.47 & 30.07 & 88.09 \\ 
    \midrule

    \multirow{3}{*}{\shortstack[l]{Llama-2-7b-\\chat-hf}} 
    & None          & 0.09 & 2.30  & 5.24 & 78.23 
    & 10k Combine   & 26.36 & 49.16  & 36.84 & 89.37 \\ 
    & 5k Known      & 19.67 & 41.86  & 31.33 & 88.06 
    & 10k Unknown   & \textbf{33.25} & \textbf{58.91} & \textbf{44.77} & \textbf{91.58} \\ 
    & 5k Unknown    & 24.57 & 48.90  & 36.69 & 89.33 
    & 10k Augmented & 21.97 & 47.57 & 35.75 & 89.14 \\ 
    \midrule

    \multirow{3}{*}{\shortstack[l]{Llama-3.1-8B-\\Instruct}} 
    & None          & 0.12 & 5.67  & 11.11 & 79.26 
    & 10k Combine   & 23.59 & 49.78  & 36.37 & 89.65 \\ 
    & 5k Known      & 14.81 & 43.14  & 31.54 & 88.50 
    & 10k Unknown   & \textbf{30.84} & \textbf{63.38} & \textbf{47.05} & \textbf{92.39} \\ 
    & 5k Unknown    & 18.96 & 49.23  & 35.95 & 89.61 
    & 10k Augmented & 22.05 & 49.56 & 36.23 & 90.12 \\ 
    \midrule

    \multirow{3}{*}{\shortstack[l]{Mistral-7B-\\Instruct-v0.1}} 
    & None          & 0.32 & 15.73  & 17.64 & 81.85 
    & 10k Combine   & 18.41 & 46.51  & 33.61 & 89.08 \\ 
    & 5k Known      & 13.20 & 39.35  & 28.41 & 87.88 
    & 10k Unknown   & \textbf{32.77} & \textbf{63.64} & \textbf{47.43} & \textbf{92.42} \\ 
    & 5k Unknown    & 15.93 & 45.52  & 32.98 & 88.92 
    & 10k Augmented & 15.67 & 45.86 & 33.21 & 89.01 \\ 
    \midrule

    \multirow{3}{*}{\shortstack[l]{Mistral-7B-\\Instruct-v0.3}} 
    & None          & 0.16 & 3.14  & 7.11 & 79.07 
    & 10k Combine   & 22.75 & 47.65  & 34.81 & 89.12 \\ 
    & 5k Known      & 16.87 & 40.87  & 29.36 & 87.94 
    & 10k Unknown   & \textbf{36.84} & \textbf{62.41} & \textbf{46.66} & \textbf{92.21} \\ 
    & 5k Unknown    & 21.21 & 47.21  & 34.42 & 89.02 
    & 10k Augmented & 16.63 & 47.13 & 34.27 & 88.99 \\ 
    \midrule

    \multirow{3}{*}{\shortstack[l]{Phi-3.5-mini-\\instruct}} 
    & None          & 0.11 & 2.41  & 5.85 & 78.86 
    & 10k Combine   & 22.93 & 43.39  & 31.60 & 89.01 \\ 
    & 5k Known      & 18.13 & 39.10  & 28.31 & 88.29 
    & 10k Unknown   & \textbf{25.72} & \textbf{47.15} & \textbf{34.19} & \textbf{89.73} \\ 
    & 5k Unknown    & 23.26 & 42.94  & 31.24 & 88.98 
    & 10k Augmented & 19.26 & 42.98 & 31.27 & 88.99 \\ 
    \midrule

    \multirow{3}{*}{\shortstack[l]{Qwen1.5-7B-\\Chat}} 
    & None          & 0.14 & 3.84  & 8.18 & 79.28 
    & 10k Combine   & 20.78 & 38.85  & 28.16 & 88.39 \\ 
    & 5k Known      & 20.04 & 39.05  & 28.53 & 88.40 
    & 10k Unknown   & \textbf{25.91} & \textbf{48.38} & \textbf{35.80} & \textbf{90.03} \\ 
    & 5k Unknown    & 19.89 & 39.02  & 28.54 & 88.40 
    & 10k Augmented & 13.85 & 38.06 & 27.56 & 88.24 \\ 
    \midrule

    \multirow{3}{*}{\shortstack[l]{Qwen2.5-7B-\\Instruct}} 
    & None          & 0.15 & 4.07  & 8.79 & 78.92 
    & 10k Combine   & 11.37 & 32.52  & 24.36 & 85.65 \\ 
    & 5k Known      & 11.72 & 32.24  & 24.18 & 85.63 
    & 10k Unknown   & \textbf{14.49} & \textbf{35.88} & \textbf{27.07} & \textbf{86.49} \\ 
    & 5k Unknown    & 11.72 & 32.30  & 24.20 & 85.63 
    & 10k Augmented & 5.82 & 31.99 & 24.24 & 85.80 \\ 
    \bottomrule
  \end{tabular}
\end{table*}

\subsection{Effectiveness of Data Augmentation}

We further investigate whether our hidden-state decoding method can generate valuable training samples to boost performance when original data is scarce. The results for the \textit{10k Augmented} setting are presented alongside the main results in the tables mentioned above.

The experiments reveal that the \textit{10k Augmented} set consistently provides a significant performance boost over the \textit{5k Unknown} set. For example, with \texttt{Phi-3.5-mini} on MedMCQA, the \textit{10k Augmented} set improves the ROUGE-L score to 29.18, compared to 27.85 for the \textit{5k Unknown} set. Notably, in several cases, the augmented data performs on par with the unfiltered \textit{10k Combine} set (e.g., 29.18 vs. 27.70 for \texttt{Phi-3.5-mini}).
This confirms RQ3: when acquiring more original "unknown" data is costly or infeasible, our hidden-state decoding method offers a practical way to enrich the training data and improve model generalization. However, it is worth noting that the performance of augmented data generally does not reach the ceiling set by real \textit{10k Unknown} data, as synthetic samples may lack the full informational depth of ground-truth data.

\subsection{Comparison with Traditional Active Learning Strategies}

To situate KA2L in the broader research context, we compared it against adapted versions of traditional active learning methods, including Entropy \citep{6889457}, Coreset \citep{sener2018active}, and BADGE \citep{ash2020deep}. 

The results, presented in Table \ref{tab:comparison_traditional_nq}, highlight a significant performance gap. On the NQ-Open dataset, KA2L (\textit{5k Unknown}) achieves a ROUGE-L score of 44.96, significantly outperforming all traditional baselines (e.g., Coreset at 38.69 and BADGE at 38.70). In fact, our method approaches the performance of the full dataset (45.18), whereas traditional methods struggle to surpass random sampling significantly.

Results on other two datasets, presented in Tables \ref{tab:comparison_traditional_medmcqa} and \ref{tab:comparison_traditional_triviaqa}, consistently corroborate this finding, demonstrating that KA2L's semantic-uncertainty-based approach is more effective at identifying informative samples for knowledge-intensive tasks than strategies based on general uncertainty, diversity, or adapted gradients.

\begin{table}[t]
\small
\centering
\caption{
    \small{\textbf{Performance comparison of active learning methods on the NQ\_Open dataset.} Our method, KA2L (\textbf{$5k$ Unknown}), significantly outperforms all adapted traditional AL method, approaching the performance of the \textbf{Full Dataset} ($10k$) with only half the data budget. Results are reported as mean ± std over $4$ runs. Bold values indicate the best performance.}
}
\label{tab:comparison_traditional_nq}
\scriptsize
\renewcommand{\arraystretch}{0.9}
\setlength{\tabcolsep}{1pt}

\begin{tabular}{l c c c c}
\toprule
\textbf{Method} & \textbf{BLEU} & \textbf{ROUGE-L} & \textbf{METEOR} & \textbf{BertScore(\%)}\\
\midrule
\multicolumn{5}{l}{\textit{Traditional Methods ($5k$ samples selected from $10k$ pool)}} \\
Random          & $21.75 \pm 0.03$ & $38.53 \pm 0.02$ & $29.65 \pm 0.02$ & $90.01 \pm 0.01$ \\
Entropy         & $19.59 \pm 1.15$ & $37.91 \pm 0.10$ & $29.38 \pm 0.07$ & $89.91 \pm 0.01$ \\
Coreset         & $22.04 \pm 0.07$ & $38.69 \pm 0.05$ & $29.81 \pm 0.03$ & $90.16 \pm 0.01$ \\
BADGE (adapted) & $19.64 \pm 0.15$ & $38.70 \pm 0.05$ & $29.73 \pm 0.05$ & $90.06 \pm 0.01$ \\
\midrule
\multicolumn{5}{l}{\textit{Our Method ($5k$ samples)}} \\
KA2L 5k Known   & $16.68 \pm 0.09$ & $32.58 \pm 0.03$ & $24.68 \pm 0.03$ & $89.07 \pm 0.01$ \\
\textbf{KA2L 5k Unknown} & $\textbf{22.29} \pm \textbf{0.08}$ & $\textbf{44.96} \pm \textbf{0.04}$ & $\textbf{35.01} \pm \textbf{0.04}$ & $\textbf{91.08} \pm \textbf{0.01}$ \\
\midrule
\multicolumn{5}{l}{\textit{Upper Bound}} \\
Full Dataset (10k) & $24.51 \pm 0.04$ & $45.18 \pm 0.04$ & $35.35 \pm 0.04$ & $91.09 \pm 0.01$ \\
\bottomrule
\end{tabular}

\end{table}

\begin{table}[t]
\small
\centering
\caption{
    \small{\textbf{Performance comparison of active learning methods on the MedMCQA dataset.} Bold values indicate the best performance.}
}
\label{tab:comparison_traditional_medmcqa}
\scriptsize
\renewcommand{\arraystretch}{0.9}
\setlength{\tabcolsep}{1pt}
\begin{tabular}{l c c c c}
\toprule
\textbf{Method} & \textbf{BLEU} & \textbf{ROUGE-L} & \textbf{METEOR} & \textbf{BertScore(\%)}\\
\midrule
\multicolumn{5}{l}{\textit{Traditional Methods (5k samples selected from 10k pool)}} \\
Random          & $7.19 \pm 0.22$ & $26.67 \pm 0.05$ & $18.55 \pm 0.05$ & $86.44 \pm 0.02$ \\
Entropy         & $6.85 \pm 0.09$ & $26.66 \pm 0.07$ & $18.46 \pm 0.05$ & $86.40 \pm 0.01$ \\
Coreset         & $8.11 \pm 0.11$ & $28.48 \pm 0.08$ & $19.86 \pm 0.03$ & $86.83 \pm 0.01$ \\
BADGE (adapted) & $6.68 \pm 0.12$ & $26.76 \pm 0.06$ & $18.64 \pm 0.07$ & $86.41 \pm 0.01$ \\
\midrule
\multicolumn{5}{l}{\textit{Our Method (5k samples)}} \\
KA2L 5k Known   & $5.67 \pm 0.06$ & $23.35 \pm 0.05$ & $16.23 \pm 0.03$ & $85.73 \pm 0.01$ \\
\textbf{KA2L 5k Unknown} & $\textbf{8.47} \pm \textbf{0.14}$ & $\textbf{29.96} \pm \textbf{0.02}$ & $\textbf{20.84} \pm \textbf{0.01}$ & $\textbf{87.19} \pm \textbf{0.01}$ \\
\midrule
\multicolumn{5}{l}{\textit{Upper Bound}} \\
Full Dataset (10k) & $8.17 \pm 0.04$ & $30.14 \pm 0.03$ & $21.08 \pm 0.02$ & $87.13 \pm 0.01$ \\
\bottomrule
\end{tabular}
\end{table}

\begin{table}[t]
\small
\centering
\caption{
    \small{\textbf{Performance comparison of active learning methods on the TriviaQA dataset.} Bold values indicate the best performance.}
}
\label{tab:comparison_traditional_triviaqa}
\scriptsize
\renewcommand{\arraystretch}{0.9}
\setlength{\tabcolsep}{1pt}
\begin{tabular}{l c c c c}
\toprule
\textbf{Method} & \textbf{BLEU} & \textbf{ROUGE-L} & \textbf{METEOR} & \textbf{BertScore(\%)}\\
\midrule
\multicolumn{5}{l}{\textit{Traditional Methods (5k samples selected from 10k pool)}} \\
Random          & $12.33 \pm 0.14$ & $45.25 \pm 0.01$ & $32.96 \pm 0.02$ & $88.95 \pm 0.01$ \\
Entropy         & $\textbf{22.34} \pm \textbf{0.08}$ & $46.25 \pm 0.09$ & $33.91 \pm 0.06$ & $89.10 \pm 0.01$ \\
Coreset         & $20.37 \pm 1.80$ & $45.64 \pm 0.07$ & $33.31 \pm 0.04$ & $89.02 \pm 0.00$ \\
BADGE (adapted) & $20.58 \pm 0.22$ & $45.50 \pm 0.06$ & $33.15 \pm 0.03$ & $88.98 \pm 0.01$ \\
\midrule
\multicolumn{5}{l}{\textit{Our Method (5k samples)}} \\
KA2L 5k Known   & $15.88 \pm 1.65$ & $43.19 \pm 0.05$ & $31.55 \pm 0.03$ & $88.51 \pm 0.01$ \\
\textbf{KA2L 5k Unknown} & $18.92 \pm 0.41$ & $\textbf{49.21} \pm \textbf{0.03}$ & $\textbf{35.93} \pm \textbf{0.02}$ & $\textbf{89.61} \pm \textbf{0.00}$ \\
\midrule
\multicolumn{5}{l}{\textit{Upper Bound}} \\
Full Dataset (10k) & $23.50 \pm 0.35$ & $49.76 \pm 0.06$ & $36.33 \pm 0.05$ & $89.65 \pm 0.01$ \\
\bottomrule
\end{tabular}
\end{table}

\section{Analysis and Discussion}
\label{sec:analysis}

While the previous section established the superior performance of KA2L, this section delves deeper into the internal mechanisms that drive its success. We analyze the behavior of the knowledge distribution probe, assess the robustness of our dynamic thresholding, and provide a qualitative visualization of knowledge boundaries. Finally, we discuss the limitations of the current framework.

\subsection{Knowledge Distribution Probing Analysis}
\label{sec:probe_analysis}

The efficacy of KA2L hinges on the probe's ability to accurately distinguish between "known" and "unknown" knowledge. To validate this, we evaluated our MLP-based probe against strong baselines (including \textit{SE Probe}, \textit{P(True)}, and \textit{Semantic Entropy}) using Area Under the ROC Curve (AUROC) as the metric.

\subsubsection{Probe Performance}
As shown in the Table \ref{tab:probe-performance-triviaqa}, our probe consistently achieves state-of-the-art performance across all evaluated LLMs. For instance, on the TriviaQA dataset, it attains an AUROC of up to 0.91 (on \texttt{Phi-3.5-mini}). Notably, our MLP-based approach significantly outperforms the linear \textit{SE Probe}, underscoring the necessity of capturing non-linear relationships in hidden states to detect subtle uncertainty signals. Similar trends are observed on MedMCQA and NQ Open.

\begin{table*}[t]
  \caption{\small{\textbf{Performance comparison of knowledge distribution probes (AUROC) on TriviaQA dataset.} Our method achieves the highest score. Bold values indicate the best performance.}}
  \label{tab:probe-performance-triviaqa}
  \scriptsize
  \renewcommand{\arraystretch}{0.9} 
  \resizebox{0.85\textwidth}{!}{ 
  \centering
    \begin{tabular}{p{2.5cm}lllllll}
    \toprule
    \multirow{2}{*}{\shortstack{\textbf{Models}}} & \multirow{2}{*}{\shortstack{\textbf{Ours}}} & \multirow{2}{*}{\shortstack{\textbf{SE Probe}}} & \multirow{2}{*}{\shortstack{\textbf{Accuracy}\\ \textbf{Probe}}} & \multirow{2}{*}{\shortstack{\textbf{Log-Likeli}\\\textbf{hood}}} & \multirow{2}{*}{\shortstack{\textbf{Regular} \\\textbf{Entropy}}} & \multirow{2}{*}{\shortstack{\textbf{P(True)}}} & \multirow{2}{*}{\shortstack{\textbf{Semantic} \\\textbf{Entropy}}} \\ \\
    
    \midrule

    DeepSeek-R1-Distill-Qwen-7B	& \multirow{2}{*}{\textbf{0.89}} 	& \multirow{2}{*}{0.85} 	& \multirow{2}{*}{0.81} 	& \multirow{2}{*}{0.61} 	& \multirow{2}{*}{0.84} 	& \multirow{2}{*}{0.83} 	& \multirow{2}{*}{0.86}  \\
    glm4-9b-chat	& \textbf{0.83} 	& 0.81 	& 0.72 	& 0.56 	& 0.76 	& 0.82 	& 0.80  \\
    Llama-2-7b-chat-hf	& \textbf{0.85} 	& 0.78 	& 0.70 	& 0.55 	& 0.73 	& 0.71 	& 0.77  \\
    Llama-3.1-8B-Instruct	& \textbf{0.89} 	& 0.85 	& 0.73 	& 0.57 	& 0.75 	& 0.79 	& 0.79  \\
    Mistral-7B-Instruct-v0.1	& \textbf{0.88} 	& 0.83 	& 0.76 	& 0.62 	& 0.75 	& 0.78 	& 0.81  \\
    Mistral-7B-Instruct-v0.3	& \textbf{0.90} 	& 0.86 	& 0.78 	& 0.68 	& 0.73 	& 0.84 	& 0.78  \\
    Phi-3.5-mini-instruct	& \textbf{0.91} 	& 0.88 	& 0.81 	& 0.77 	& 0.82 	& 0.79 	& 0.84  \\
    Qwen1.5-7B-Chat	& \textbf{0.81} 	& 0.78 	& 0.74 	& 0.46 	& 0.77 	& 0.78 	& \textbf{0.81}  \\
    Qwen2.5-7B-Instruct	& \textbf{0.86} 	& 0.82 	& 0.75 	& 0.54 	& 0.79 	& \textbf{0.86} 	& 0.80  \\

  \bottomrule
    \end{tabular}
    }
\end{table*}

\subsubsection{Layer-wise Analysis}
\label{sec:layer-wise-analysis}

\begin{figure*}[t]
    \centering
    \includegraphics[width=0.90\linewidth]{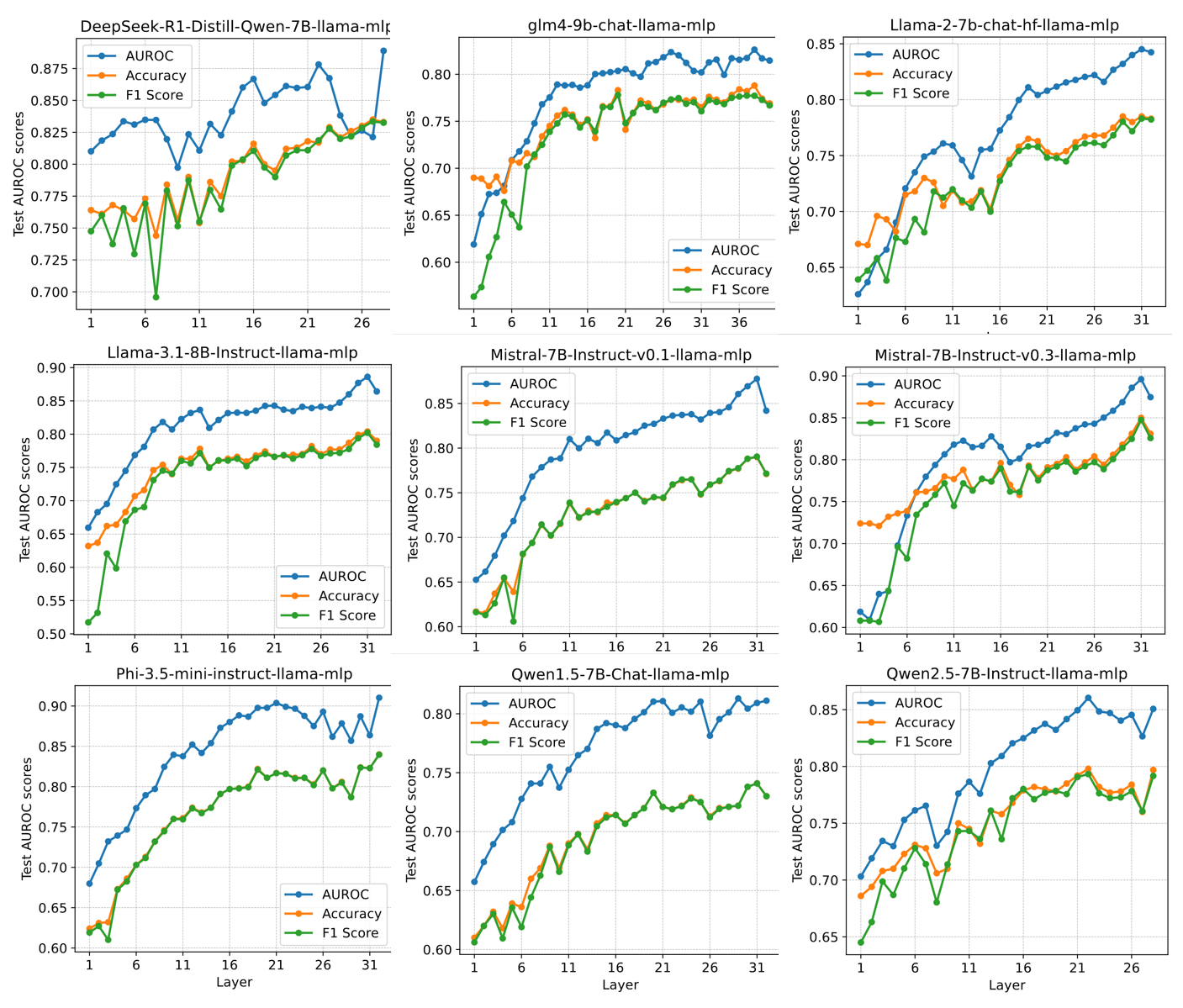}
    \caption{\small{\textbf{Classifier AUROC score across different hidden layers} (embedding layer excluded) on the Trivia\_QA dataset.}}
    \label{fig:layer_analysis}
\end{figure*}

To determine the optimal source of uncertainty signals, we analyzed the classifier performance across different Transformer layers. Figure \ref{fig:layer_analysis} visualizes the AUROC scores across layers for different models.

We observed a consistent trend: classifier performance generally improves with increasing layer depth, suggesting that deeper layers contain more semantic-level information relevant to output uncertainty. However, there are distinct architectural signatures. For example, \texttt{Llama-3.1-8B} shows a steady rise, whereas \texttt{Phi-3.5-mini} plateaus earlier. This analysis not only guided our layer selection in Section \ref{sec:exp_design} but also confirms that knowledge uncertainty is encoded deeply within the model's high-level representations.

\subsection{Robustness of Dynamic Thresholding}
\label{sec:threshold_robustness}

Our framework utilizes a dynamic thresholding mechanism (minimizing MSE) to binarize Semantic Entropy labels. To assess whether our method is overly sensitive to this hyperparameter, we conducted a perturbation analysis on \texttt{Llama-3.1-8B}. We perturbed the optimal threshold $\gamma^*$ by increments of $\pm 0.05, \pm 0.10, \pm 0.20$ and re-evaluated the probe's accuracy.

The results shown in Table \ref{tab:threshold_robustness} demonstrate that the classifier consistently achieves the highest AUROC score precisely at the optimal threshold \(\gamma^*\) across all three datasets. This empirically validates that our MSE-based method is effective at identifying an optimal cut-off point. Furthermore, the performance degrades gracefully as the threshold deviates from the optimum. Even with a significant perturbation, the performance does not collapse, indicating that our framework is not overly sensitive to the exact threshold value and demonstrates practical robustness.

\begin{table*}[t]
\centering
\scriptsize
\caption{AUROC scores of the knowledge distribution probe under perturbed binarization thresholds for \texttt{LLaMA-3.1-8B-Instruct} (Layer $31$). The optimal threshold \(\gamma^*\) is determined by minimizing MSE. The highest score in each row is highlighted in bold.}
\label{tab:threshold_robustness}
\begin{tabular}{lccccccc}
\toprule
\textbf{Dataset} & \textbf{\(\gamma^*-0.20\)} & \textbf{\(\gamma^*-0.10\)} & \textbf{\(\gamma^*-0.05\)} & \textbf{\(\gamma^*\)} & \textbf{\(\gamma^*+0.05\)} & \textbf{\(\gamma^*+0.10\)} & \textbf{\(\gamma^*+0.20\)} \\
\midrule
TriviaQA & 0.8543 & 0.8602 & 0.8623 & \textbf{0.8853} & 0.8847 & 0.8784 & 0.8721 \\
NQ\_Open & 0.8232 & 0.8290 & 0.8277 & \textbf{0.8307} & 0.8265 & 0.8248 & 0.8247 \\
MedMCQA & 0.8281 & 0.8267 & 0.8323 & \textbf{0.8345} & 0.8285 & 0.8238 & 0.8333 \\
\bottomrule
\end{tabular}
\end{table*}

\subsection{Qualitative Analysis: Visualizing Knowledge Boundaries}
\label{sec:qualitative_analysis}
To provide an intuitive understanding of what KA2L considers "Known", "Unknown" and the "Knowledge Boundary", we present a qualitative case study in Table \ref{tab:refined_case_study}. This table details the generative distribution of 10 independent samples for questions across different SE ranges.

\begin{figure}
\centering
\includegraphics[width=0.60\linewidth]{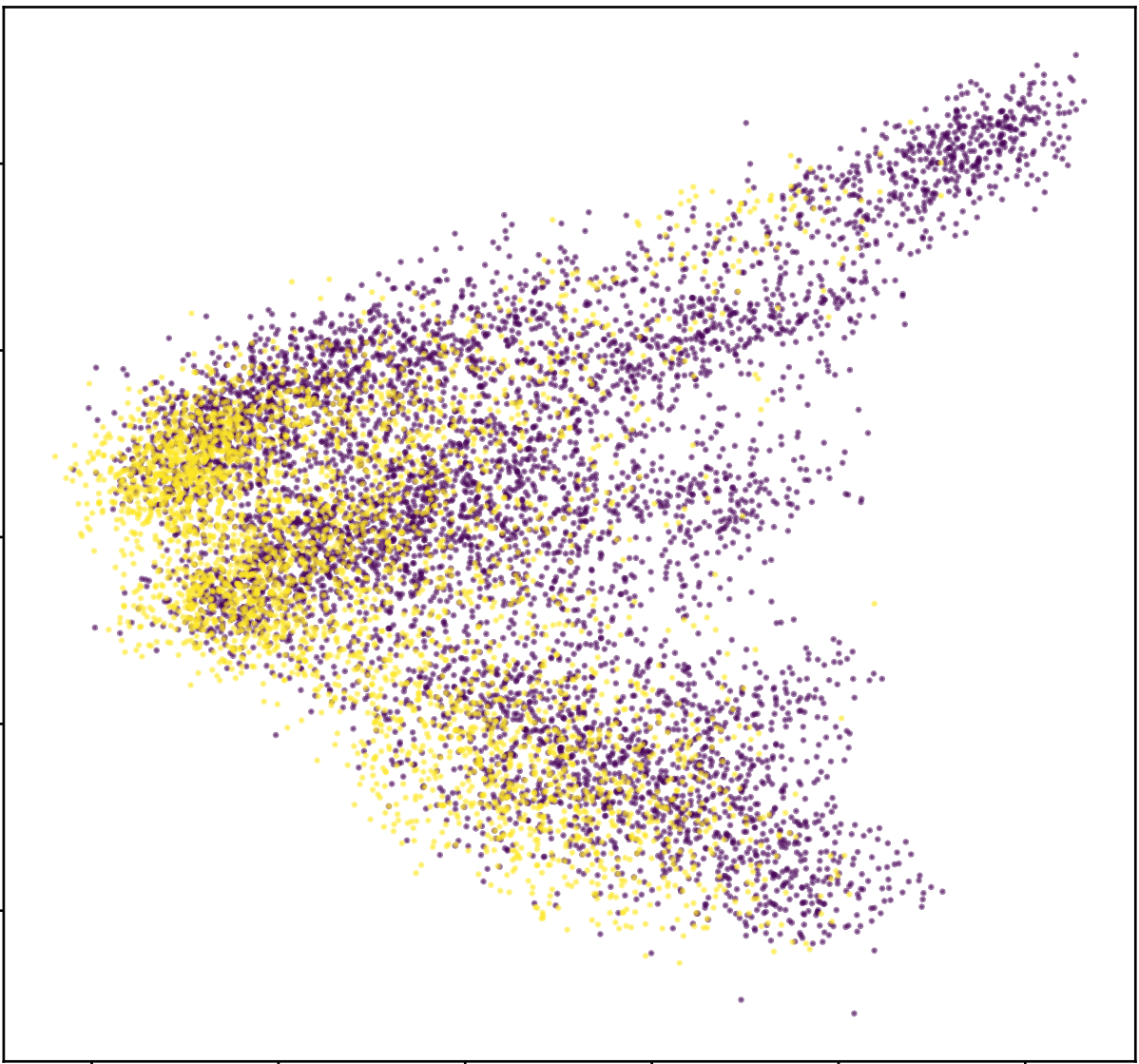}
\caption{PCA Analysis on Mistral-7B-Instruct-v0.3, showing a macroscopic overview of knowledge distribution.}
\label{fig:pca_plot}
\end{figure}

\textbf{Low Entropy (Known):} For questions with SE $\approx 0$, the model consistently generates the correct factual core across all 10 samples. Minor lexical variations do not alter the semantic meaning, indicating confidently mastered knowledge which KA2L correctly filters out to avoid redundant training.
    
\textbf{Medium Entropy (The Knowledge Boundary):} The intermediate cases perfectly visualize the epistemic "knowledge boundary". For example, when asked about the origin of Häagen-Dazs, the model generates a superposition of the correct answer ("United States") and plausible hallucinations ("Poland", "Scandinavia"). This semantic instability demonstrates the model's uncertainty at the very edge of its parametric knowledge.
    
\textbf{High Entropy (Unknown):} For high SE questions ($\approx 2.3$), the 10 independent generations diverge chaotically into various incorrect entities or nonsensical phrases, proving a complete lack of parametric knowledge and validating their selection as high-value targets for active learning.

\begin{table*}[htbp]
\centering
\renewcommand{\arraystretch}{0.6} 
\footnotesize
\caption{Case Study: Generative distributions of 10 independent samples across different SE ranges. The Medium SE category explicitly visualizes the "Knowledge Boundary" where outputs mix correct facts with plausible hallucinations.}
\label{tab:refined_case_study}
\resizebox{0.9\textwidth}{!}{
\begin{tabular}{p{0.02\textwidth} p{0.28\textwidth} p{0.15\textwidth} p{0.48\textwidth}}
\toprule
\textbf{\#} & \textbf{Question} & \textbf{Ground Truth} & \textbf{Generated Answers (10 Samples)} \\
\midrule
\multicolumn{4}{c}{\textbf{Low Entropy (Known Knowledge, SE $\approx$ 0)}} \\
\midrule
1 & how many keys are on microsoft windows standard keyboard & 104 & \texttt{104} (10 times) \\
2 & who played willy wonka in the original movie & Gene Wilder & \texttt{Gene Wilder} (8 times), \texttt{gene wilder} (2 times) \\
3 & how many original harry potter books are there & 7 & \texttt{7} (6 times), \texttt{Seven} (2 times), \texttt{seven} (2 times) \\
\midrule
\multicolumn{4}{c}{\textbf{Medium Entropy (The Knowledge Boundary, SE $\approx$ 1.3 - 1.7)}} \\
\midrule
4 & the haagen-dazs brand of ice cream originated in which country & United States & \makecell[l]{\texttt{United States} (2), \texttt{Poland} (4), \texttt{Scandinavia} (1),\\ \texttt{scandinavian} (1), \texttt{Belgium} (1), \texttt{Jamaica} (1)} \\
5 & who has the most superbowl wins in nfl history & Pittsburgh Steelers & \makecell[l]{\texttt{Pittsburgh Steelers} [incl. variations] (7),\\ \texttt{Tom Brady} (2), \texttt{Green Bay Packers} (1)} \\
6 & what are the main components of earth's atmosphere & nitrogen & \makecell[l]{\texttt{Nitrogen, Oxygen, and Argon} [incl. variations] (9),\\ \texttt{Nitrogen Oxygen Carbon...} (1)} \\
\midrule
\multicolumn{4}{c}{\textbf{High Entropy (Unknown Knowledge, SE $\approx$ 2.3)}} \\
\midrule
7 & who holds the record for most knockouts in boxing & Archie Moore & \makecell[l]{\texttt{Rocky Marciano}, \texttt{Chris Byrd}, \texttt{Muhammad Ali}, \texttt{Jabbar Ali},\\ \texttt{Tyson}, \texttt{Joe Louis} (2), \texttt{Partinello Wolf}, \texttt{George Foreman}, \texttt{Yvon Neptune}} \\
8 & who is the actual president of the supreme court of justice & John G. Roberts & \makecell[l]{\texttt{Joy Bivoni}, \texttt{kennedy library}, \texttt{Justice enrique}, \texttt{María Adelina Díaz},\\ \texttt{Silvia Bercioux}, \texttt{Juan Cueto}, \texttt{Enrique Iglesias}, \texttt{Problem unknown}, etc.} \\
9 & what is the average salary of a working american & \$59,039 & \makecell[l]{\texttt{Per computronics 71,301}, \texttt{\$47,940}, \texttt{\$58,715}, \texttt{\$54,000}, \texttt{\$55,292},\\ \texttt{\$48,232.42}, \texttt{\$74841 USD}, \texttt{\$50,708 in May 2021}, \texttt{\$64,968}, \texttt{\$46,641}} \\
\bottomrule
\end{tabular}
}
\end{table*}

We also performed PCA analysis on the 10k Combine dataset and \texttt{Mistral-7B-Instruct-v0.3} model based on Semantic Entropy values. The macroscopic projection in Figure \ref{fig:pca_plot} visualizes a structural overlapping area between the "Known" and "Unknown" distributions, conceptually representing the knowledge boundary. Complementing this macro-level visualization, the multi-sampling generative distributions in Table \ref{tab:refined_case_study} provide fine-grained qualitative evidence, concretely illustrating how the model's epistemic uncertainty manifests as semantic instability at these boundaries.

\subsection{Intrinsic Analysis of Hidden-State Decoding}
\label{sec:intrinsic_analysis}

To further validate the efficacy of the T5-base hidden-state decoder, we conducted a comprehensive intrinsic analysis on the generated question variants. Following the default configuration of \citet{morris2024language}, we retained the T5-base backbone, which offers an optimal trade-off between generation quality and the massive computational efficiency required for active learning. Specifically, for each dataset, we randomly sampled 1,000 questions identified as "unknown" from the \texttt{Llama-3.1-8B-Instruct} model, and generated $k=5$ candidate variants per question to systematically evaluate the decoder's performance.

We performed both quantitative and qualitative evaluations across three datasets, as shown in Table \ref{tab:intrinsic_metrics} and Table \ref{tab:case_study_variants}. Quantitatively, the exact-copy rate across all datasets is 0.0, indicating that the decoder does not merely memorize or duplicate inputs. Furthermore, out of the 5 generated candidates, it successfully produces multiple unique, valid variants per source question (averaging 3.19 to 3.51 usable variants after deduplication). These generated variants maintain a high degree of semantic relatedness to the original questions, with the Source BERTScore F1 ranging from 0.880 to 0.913.

\begin{table}[htbp]
\centering
\caption{Intrinsic quantitative evaluation of generated question variants. \textit{Avg Usable Variants} indicates the ratio of unique variants among 5 generations. \textit{Source BERTScore} measures semantic relatedness to the original question. \textit{Self-BLEU} measures the lexical diversity among the variants themselves.}
\label{tab:intrinsic_metrics}
\resizebox{\columnwidth}{!}{
\begin{tabular}{lcccc}
\toprule
\textbf{Dataset} & \textbf{Avg Usable Variants} & \textbf{Source BERTScore} & \textbf{Self-BLEU} & \textbf{Exact-Copy Rate} \\
\midrule
MedMCQA & 3.51 & 0.908 & 42.24 & <0.01 \\
NQ\_Open & 3.37 & 0.880 & 35.01 & <0.01 \\
TriviaQA & 3.19 & 0.913 & 48.13 & <0.01 \\
\bottomrule
\end{tabular}
}
\end{table}

Qualitatively, the generated variants exhibit characteristics that are highly beneficial for data augmentation. In active learning and supervised fine-tuning, producing identically phrased questions offers limited value. As illustrated in Table \ref{tab:case_study_variants}, our decoder introduces valuable \textit{exploratory diversity} rather than simple paraphrasing. For instance, while it preserves the structural integrity of the questions in TriviaQA, it frequently performs \textit{entity substitution} in NQ\_Open (substituting one entity for another within the same domain) and \textit{terminological perturbation} in MedMCQA. Because these exploratory variants often introduce new entities or shifted factual premises, they are treated as independent queries and re-annotated with their own correct answers during downstream processing. Generating these parallel questions within the same knowledge neighborhood prevents the LLM from overfitting to specific entities, effectively broadening the distribution of the augmented data and promoting better generalization.

\begin{table*}[htbp]
\centering
\caption{Qualitative examples of generated question variants. The decoder introduces beneficial diversity through structural rephrasing, entity substitution, and terminology perturbation, effectively broadening the augmented data distribution.}
\label{tab:case_study_variants}
\resizebox{0.85\textwidth}{!}{
\begin{tabular}{p{0.18\textwidth} p{0.38\textwidth} p{0.55\textwidth}}
\toprule
\textbf{Augmentation Trait} & \textbf{Original Source Question} & \textbf{Generated Question Variants} \\
\midrule
\textbf{Structural Rephrasing} (TriviaQA) & Ex-world champion boxer Alan Rudkin, who died in September 2010, fought at what weight? & \textbullet~Ex-world champion boxer Alan Rudkin, who died in September 2010, was fighting at what weight and age in 1960s? \\
\midrule
\textbf{Entity Substitution} (NQ\_Open) & Which player has won the most UEFA champions league titles? & \textbullet~Which player has won the most trophies in the European cup 2018? \newline \textbullet~Which team has won the most trophies in the European cup 2017? \\
\midrule
\textbf{Terminological Perturbation} (MedMCQA) & Type of immunologic injury in AIHA (Auto immune hemolytic anemia) & \textbullet~Type of immunological response to IgA hemolytic anemia (IgA) is \\
\bottomrule
\end{tabular}
}
\end{table*}

\vspace{1em}
\noindent\textbf{Impact of Training Data Composition.} 
We further investigate the effect of the general-to-domain data ratio on the decoder's performance. To isolate the impact of data composition from the total training volume, we fixed the training set size at 250,000 examples and varied the domain-specific proportion. We evaluated the generation metrics using \texttt{Llama-3.1-8B-Instruct} on 1,000 held-out questions per dataset (Table \ref{tab:ratio_ablation}).

As shown in the results, incorporating domain-specific data consistently improves semantic relatedness (BERTScore) compared to the pure general-domain baseline. However, increasing the domain ratio beyond approximately 20\% yields diminishing returns, with metrics plateauing or slightly fluctuating on NQ\_Open and MedMCQA. Furthermore, the Exact-Copy Rate remains near zero across all settings, indicating that the decoder effectively learns the domain distribution without merely memorizing inputs. In practice, acquiring massive amounts of high-quality, domain-specific data is often expensive and challenging. These empirical results suggest that an approximately 20\% domain data ratio serves as a highly cost-effective heuristic. It captures the majority of domain-adaptation benefits, whereas pursuing higher ratios entails escalating data collection costs for only marginal or fluctuating performance gains. This validates our 200k-general/50k-domain configuration as a practical and efficient operating point for hidden-state decoding.

\begin{table*}[h]
\centering
\caption{Effect of decoder-training data composition on generation metrics using \texttt{Llama-3.1-8B-Instruct}. The total training size is fixed at 250,000. Bold values indicate the highest metric among the evaluated ratios.}
\label{tab:ratio_ablation}
\resizebox{0.95\textwidth}{!}{
\begin{tabular}{lrrrrrrrr}
\toprule
\textbf{Dataset} & \textbf{Domain Ratio} & \textbf{General} & \textbf{Domain} & \textbf{BLEU} & \textbf{ROUGE-L} & \textbf{METEOR} & \textbf{BERTScore} & \textbf{Exact-Copy Rate} \\ \midrule
NQ\_Open & 0\% & 250,000 & 0 & 5.42 & 22.94 & 39.18 & 85.83 & <0.01 \\
NQ\_Open & 10\% & 225,000 & 25,000 & 6.24 & 23.39 & 41.92 & 86.40 & <0.01 \\
NQ\_Open & 20\% & 200,000 & 50,000 & \textbf{7.67} & \textbf{26.79} & \textbf{45.98} & \textbf{87.16} & <0.01 \\
NQ\_Open & 30\% & 175,000 & 75,000 & 6.94 & 25.37 & 44.07 & 86.93 & <0.01 \\ \midrule
TriviaQA & 0\% & 250,000 & 0 & 16.12 & 43.56 & 50.33 & 89.57 & <0.01 \\
TriviaQA & 10\% & 225,000 & 25,000 & 19.16 & 47.07 & 53.18 & 90.22 & <0.01 \\ 
TriviaQA & 20\% & 200,000 & 50,000 & 20.37 & 48.18 & 55.24 & 90.50 & <0.01 \\
TriviaQA & 30\% & 175,000 & 75,000 & \textbf{24.32} & \textbf{52.57} & \textbf{58.94} & \textbf{91.14} & <0.01 \\ \midrule
MedMCQA & 0\% & 250,000 & 0 & 10.47 & 34.28 & 35.60 & 86.42 & <0.01 \\
MedMCQA & 10\% & 225,000 & 25,000 & 21.11 & 49.96 & 51.48 & 89.34 & 0.01 \\ 
MedMCQA & 20\% & 200,000 & 50,000 & 27.29 & 55.98 & 56.29 & 90.33 & 0.01 \\
MedMCQA & 30\% & 175,000 & 75,000 & \textbf{30.71} & \textbf{60.02} & 60.85 & 91.23 & 0.01 \\
MedMCQA & 40\% & 150,000 & 100,000 & 26.32 & 56.45 & 60.24 & 90.85 & <0.01 \\
MedMCQA & 50\% & 125,000 & 125,000 & 28.54 & 58.55 & \textbf{62.21} & \textbf{91.24} & <0.01 \\ \bottomrule
\end{tabular}
}
\end{table*}

\subsection{Computational Cost-Efficiency Analysis}
\label{sec:computational_cost}

All experiments in this paper were conducted on NVIDIA A100 40GB GPUs. All GPU hours mentioned below are based on the usage of this GPU model. 

To provide a practical perspective for practitioners, we first outline the computational cost for a single model-dataset pair. The process includes:

\begin{itemize}
    \item \textbf{Probe Training and Inference:} Although we train independent MLP classifiers for all $L$ hidden layers to identify the optimal layer, the MLP architecture is extremely lightweight. Consequently, training all $L$ classifiers on 10,000 samples and performing inference requires a total of only 5 A100-GPU hours. The inference time of the trained probe is negligible.
    \item \textbf{Decoder Training:} Training the T5-based hidden-state decoder on roughly 250k samples takes about 20 A100-GPU hours. Its inference is also highly efficient and parallelizable.
\end{itemize}

It is crucial to interpret these figures (approximately 25 GPU hours in total) as a one-time, upfront investment for a given use case. In practice, 25 GPU hours represents a minimal financial cost compared to the immense human labor and time required for domain-specific data annotation. This initial cost yields a significant return by saving approximately 50\% in subsequent annotation and fine-tuning costs, which are consistently the most expensive bottlenecks in the LLM development cycle. This empirically substantiates the overall cost-effectiveness of our framework.

The total computational budget for the experiments in this paper was approximately 900 GPU hours. It is important to note that this figure reflects the cost of a comprehensive scientific validation process designed to test our framework's robustness across diverse conditions, rather than the cost of a single practical application. This large-scale effort, spanning 9 LLMs and 3 datasets, can be broken down as follows:
\begin{itemize}
    \item A total of 351 GPU hours were dedicated to evaluating the active learning process. This involved training 27 distinct knowledge probes, followed by 135 fine-tuning runs and 162 evaluation runs to compare different data selection strategies.
    \item A total of 540 GPU hours were allocated to validate the data augmentation component. The majority of this time was spent training the 27 hidden-state decoders required for our analysis.
\end{itemize}

\subsection{Limitations}
\label{sec:limitations}

While our KA2L framework effectively identifies questions representing unmastered knowledge to guide SFT dataset construction, its current functionality has several limitations. Firstly, the framework's scope is confined to question classification and prioritization; it does not extend to the automatic generation of complete question-answer pairs for the identified unknown questions, a step that still necessitates external mechanisms or human annotation. Secondly, as a white-box approach, KA2L requires access to the model's internal hidden states, rendering it incompatible with commercial, closed-source models accessible only via APIs. Lastly, the initiation of the KA2L framework is contingent upon a pre-existing, domain-specific question set to probe the LLM's knowledge distribution. The collection and curation of this initial, comprehensive corpus can pose a significant practical challenge, particularly in niche or emerging domains where such resources are scarce, thereby potentially limiting the immediate applicability or bootstrapping of our method in these scenarios. Finally, although we validated our framework on a highly specialized medical dataset (MedMCQA), extending the evaluation of KA2L to other vertical domains (e.g., law or finance) remains a promising direction for future work.

\section{Conclusion and Future Works}

In this paper, we introduced the Knowledge-Aware Active Learning (KA2L) framework, a novel approach for efficiently fine-tuning Large Language Models. By probing the model's internal hidden states to identify ``unknown" knowledge, KA2L guides a more targeted and cost-effective data selection process. Our extensive experiments demonstrate that fine-tuning with KA2L-selected data not only reduces annotation and computation costs by approximately $50\%$ but also achieves superior performance compared to both unfiltered datasets and classic active learning methods like Coreset and BADGE. The core of our framework, a highly accurate knowledge probe, effectively pinpoints a model's knowledge boundaries, while our hidden-state decoding offers a practical solution for data augmentation in low-resource scenarios. Ultimately, KA2L presents a practical and robust solution for targeted LLM enhancement, highlighting the significant value of leveraging a model's internal knowledge distribution for more efficient learning.

Our work leads to two main future research directions. A key challenge that warrants further investigation is the inherent difficulty for semantic consistency-based methods, including Semantic Entropy, to fully disambiguate between genuine knowledge gaps (i.e., the model truly does not know) and responses to questions with intrinsic ambiguity or controversy. Although our current study mitigates this issue by focusing on factual question-answering datasets, addressing this distinction in more general and open-ended scenarios remains a significant open problem. Another promising direction involves moving beyond the empirical selection of the optimal layer for uncertainty quantification. Future work could delve into the information flow mechanisms within diverse LLM architectures to develop a more principled understanding of why specific layers are more sensitive to representing knowledge uncertainty. This line of inquiry not only promises to enhance the robustness and theoretical grounding of our framework but also contributes valuable insights to the broader field of model interpretability.

\section*{Acknowledgments}
This study was supported in part by a grant from the National Key Research and Development Program of China [2025YFE0209200], the Key R\&D Program of Heilongjiang Province (China)[2024ZX01A07], and the Key R\&D Program of Heilongjiang Province (China)[JD2023GJ01].

\printcredits

\section*{Declaration of Generative AI Usage}
During the preparation of this work the authors used \texttt{Gemini-2.5-pro} in order to assist with language translation and polishing to improve grammatical clarity. After using this tool/service, the authors reviewed and edited the content as needed and take full responsibility for the content of the published article.

\section*{Declaration of competing interests}
The authors declare that they have no known competing financial interests or personal relationships that could have appeared to influence the work reported in this paper.

\bibliographystyle{apalike}

\bibliography{cas-refs}

\end{document}